\documentclass{article}
\usepackage[utf8]{inputenc}
\usepackage{amsmath}
\usepackage{amsfonts}
\usepackage{amssymb}
\usepackage{graphics}
\usepackage{graphicx}
\usepackage{caption}
\usepackage{subcaption}

\title{A Tutorial on VAEs: From Bayes' Rule to Lossless Compression}
\author{Ronald Yu\\ ronaldyu@ucsd.edu }
\date{June 2020}

\begin{document}

\maketitle

\begin{abstract}
    The Variational Auto-Encoder (VAE) is a simple, efficient, and popular deep maximum likelihood model. 
    Though usage of VAEs is widespread, the derivation of the VAE is not as widely understood.
    In this tutorial, we will provide an overview of the VAE and a tour through various derivations and interpretations of the VAE objective. From a probabilistic standpoint, we will examine the VAE through the lens of Bayes' Rule, importance sampling, and the change-of-variables formula. From an information theoretic standpoint, we will examine the VAE through the lens of lossless compression and transmission through a noisy channel. 
    We will then identify two common misconceptions over the VAE formulation and their practical consequences.
    Finally, we will visualize the capabilities and limitations of VAEs using a code example (with an accompanying Jupyter notebook) on toy 2D data.
\end{abstract}

\section{Introduction}

The Variational Auto-Encoder (VAE) belongs to a class of models, which we will refer to as \emph{deep maximum likelihood models}, that uses a deep neural network to learn a maximum likelihood model for some input data.
They are perhaps the most simple and efficient deep maximum likelihood model available, and have thus gained popularity in representation learning and generative image modeling.
Unfortunately, in my opinion, in some circles the term ``VAE" has become somewhat synonymous with ``an auto-encoder with stochastic regularization that generates useful or beautiful samples", which has led to various misconceptions about VAEs.
In this tutorial, we will return to the probabilistic and information theoretic roots of VAEs, clarify common misconceptions about VAEs, and look at a toy example on 2D data that will illustrate the capabilities and limitations of VAEs.

In Section~\ref{sec:overview}, we will give an overview of what is a maximum likelihood model and what a VAE looks like.

In Sections~\ref{sec:prob} and~\ref{sec:information}, we will motivate the VAE and obtain intuitive insight into its behavior by deriving its objective.
This derivation is broken into a probabilistic interpretation (Section~\ref{sec:prob})---in which we view a VAE through the lens of Bayes' Rule, importance sampling, and the change of variables formula---and an information theoretic interpretation(Section~\ref{sec:information})---in which we view a VAE through the lens of lossless compression and transmission through a noisy channel.

In Section~\ref{sec:misconceptions}, we will clarify two misconceptions about VAEs that I have encountered in casual conversation and teaching materials: that they can be trained using the mean-squared error loss and that the latent vector of the VAE can be viewed as a parameter rather than a variable. These two misconceptions over the formulation may lead to the incorrect beliefs that VAEs have blurry reconstructions or that they can only model Gaussian data.

Finally, in Section~\ref{sec:code}, we will gain insight into the capabilities and limitations of VAEs through a code example on toy 2D data. In this code example, we will visualize the VAE's density estimation abilities and latent space. An accompanying Jupyter Notebook is provided.

\section{VAE Overview}
\label{sec:overview}
\subsection{What is Maximum Likelihood?}
Suppose we have high-dimensional data that follows a ground truth distribution $p_{gt}(\mathbf{x})$.
A maximum likelihood model learns a probabilistic model $p_\theta(\mathbf{x})$ parameterized by $\theta$ that seeks to approximate $p_{gt}(\mathbf{x})$.
We can do so by collecting i.i.d. samples from $p_{gt}(\mathbf{x})$ to create a training set $\mathcal{D}=\{\mathbf{x}^{(1)}, \dots, \mathbf{x}^{(N)} \}$, and learning to maximize the likelihood of the joint distribution $p_\theta(\mathbf{x}^{(1)}, \dots, \mathbf{x}^{(N)}) = \prod_{i=1}^N p_\theta(\mathbf{x}^{(i)})$. For numerical stability, we instead minimize the negative log-likelihood:
\begin{align}
    -\log p_\theta(\mathbf{x}^{(1)}, \dots, \mathbf{x}^{(N)})
    = \sum_{i=1}^N -\log p_\theta(\mathbf{x}^{(i)}) \label{eq:empirical}
\end{align}
As is the case with virtually all machine learning models, we hope that by minimizing the empirical risk of our training set given by Equation~\ref{eq:empirical}, we will also minimize the true risk $\mathbb{E}_{\mathbf{x}\sim p_{gt}(\mathbf{x})} -\log p_\theta(\mathbf{x})$, which reaches a global minima if and only if $p_{gt}(\mathbf{x}) = p_\theta(\mathbf{x})$.

Two key operations for a maximum likelihood model are \emph{inference} and \emph{generation}. Inference is the ability to evaluate $p_\theta(\mathbf{x})$ for any input vector $\mathbf{x}\in \mathbb{R}^n$. Generation is the ability to sample data  from the distribution  $p_\theta(\mathbf{x})$.
In the asymptotic case
where $p_\theta(\mathbf{x})$ approaches $p_{gt}(\mathbf{x})$, one application for inference is out-of-distribution data detection (e.g. adversarial examples).
One application for generation is generative image modeling.
However, existing maximum likelihood models are currently not powerful enough to reliably perform out-of-distribution detection or sample images that come close to achieving the diversity and perceptual quality of natural images.

\subsection{What is a VAE?}

The VAE performs inference and generation by introducing a latent variable $\mathbf{z}$ that follows a \emph{prior} distribution $p_\theta(\mathbf{z})$. The VAE then uses an auto-encoder with an encoder parameterized by $\phi$ and a decoder parameterized by  $\theta$ to infer a \emph{posterior} distribution $q_\phi(\mathbf{z}|\mathbf{x})$ and an output distribution  $p_\theta(\mathbf{x}|\mathbf{z})$. If inference and generation can be efficiently done for all three of these distributions, then they can also be done on $p_\theta(\mathbf{x})$.
As in the case of a standard auto-encoder, the decoder tries to reconstruct the input $\mathbf{x}$ given a latent variable $\mathbf{z}$. The encoder predicts which $\mathbf{z}$ would be most capable of reconstructing $\mathbf{x}$.
One advantage of latent models such as VAEs over maximum likelihood models without latent variables is the potential application of $\mathbf{z}$ for semi-supervised or disentangled representation learning.

During generation, a latent vector $\mathbf{z}$ is sampled from $p_\theta(\mathbf{z})$, and the decoder outputs the parameters of $p_\theta(\mathbf{x}| \mathbf{z})$, from which we can sample an output vector.
While exact inference cannot typically be efficiently done using a VAE, we can efficiently estimate an \emph{upper-bound} of the negative log-likelihood $\log p_\theta(\mathbf{x})$ given by:
\begin{align}
    -\log p_\theta(\mathbf{x}) \leq  \mathbb{E}_{\mathbf{z} \sim  q_\phi(\mathbf{z}|\mathbf{x})}[- \log p_\theta(\mathbf{x}| \mathbf{z})] + D_{KL}(q_\phi(\mathbf{z}|\mathbf{x}) || p_\theta(\mathbf{z})) \label{eq:elbo}
\end{align}
where $D_{KL}$ refers to the KL-Divergence. 
This bound is also commonly referred to as the \emph{negative Evidence Lower-BOund (ELBO)}, and can be denoted as $-\mathcal{L}(\mathbf{x})$.
During training, we can use the negative ELBO as an objective, which in turn minimizes $-\log p_\theta(\mathbf{x})$. 
Before deriving Equation~\ref{eq:elbo} and discussing its intuitive meaning, let us first give a concrete example of what a VAE could look like in terms of neural network outputs.

\subsection{A Typical VAE}
\label{sec:example}
The prior distribution is typically a standard isotropic multi-variate Gaussian $p_\theta(\mathbf{z})=\mathcal{N}(0,\mathbf{I})$.
The inferred posterior distribution is typically a multi-variate Gaussian with diagonal co-variance $q_\phi(\mathbf{z}|\mathbf{x}) = \mathcal{N}(\boldsymbol \mu_\phi(\mathbf{x}), diag(\boldsymbol \sigma^2_\phi (\mathbf{x}))$.
Practically speaking, our encoder would be a deep neural network that consumes $\mathbf{x}$ as input and outputs two vectors $\boldsymbol \mu_\phi(\mathbf{x}), \boldsymbol \sigma^2_\phi (\mathbf{x})\in \mathbb{R}^d$ where $d$ is the dimensionality of the latent space.

Sampling  $\mathbf{z} \sim q_\phi(\mathbf{z}|\mathbf{x})$ can be done using the \emph{reparameterization trick}, by first sampling a random vector $\mathbf{u} \sim \mathcal{N}(0,\mathbf{I})$ and letting 
\begin{align}
    \mathbf{z} = \boldsymbol \mu_\phi(\mathbf{x}) + \boldsymbol \sigma_\phi(\mathbf{x}) \odot \mathbf{u} \label{eq:reparameterize}
\end{align}
where $\odot$ is the element-wise product. Since $\mathbf{z}$ is a deterministic function of  $ \boldsymbol \mu_\phi(\mathbf{x})$ and $\sigma^2_\phi(\mathbf{x})$, the resulting gradient with respect to the encoder is quite stable.

There is more variety for how the output distribution $p_\theta(\mathbf{x}|\mathbf{z})$ can be parameterized. One solution is to use an isotropic multi-variate Gaussian $p_\theta(\mathbf{x}|\mathbf{z}) = \mathcal{N}(\boldsymbol \mu_\theta(\mathbf{z}), diag(\boldsymbol \sigma^2_\theta(\mathbf{z})) )$.
Practically speaking, our decoder would be a deep neural network that consumes $\mathbf{z}$ as input and outputs the vectors 
$\boldsymbol \mu_\theta(\mathbf{z}), \boldsymbol \sigma^2_\theta(\mathbf{z}) \in \mathbb{R}^n$.

We are now ready to discuss the objective in terms of neural network outputs.
Under this formulation, the first term in Equation~\ref{eq:elbo}, which is also commonly referred to as the \emph{reconstruction loss} or $L_{rec}$, is given by:
\begin{align}
    L_{rec} &= \mathbb{E}_{\mathbf{z} \sim  q_\phi(\mathbf{z}|\mathbf{x})}[- \log p_\theta(\mathbf{x}| \mathbf{z})] \\
     &= \mathbb{E}_{\mathbf{z} \sim  q_\phi(\mathbf{z}|\mathbf{x})}[
     \sum_{i=1}^n \frac{1}{2}\log 2\pi \sigma^2_\theta(\mathbf{z})_i + \frac{(x_i - \mu_\theta(\mathbf{z})_i)^2}{2\sigma^2_\theta(\mathbf{z})_i}]
     \label{eq:reconloss}
\end{align}
where $v_i$ indicates the $i$th element of a vector $\mathbf{v}$.
$L_{rec}$ is approximated via Monte-Carlo sampling; however, due to computational constraints, during training $\mathbf{z}$ is typically only sampled once per iteration.

Under our formulation, the second term in Equation~\ref{eq:elbo}, which is also commonly referred to as the \emph{regularization loss} or $L_{reg}$, has a closed form expression given by:
\begin{align}
     L_{reg} &=  D_{KL}(q_\phi(\mathbf{z}|\mathbf{x}) || p_\theta(\mathbf{z})) \\
     &= \frac{1}{2} ||\boldsymbol \mu_\phi(\mathbf{x})||_2^2 + \frac{1}{2} [ ||\boldsymbol \sigma_\phi(\mathbf{x}) ||_2^2 - d - \sum_{i=1}^d \log \sigma^2_{\phi}(\mathbf{x})_i] \label{eq:regloss}
\end{align}

The decoder is only affected by the reconstruction loss and seeks to best reconstruct $\mathbf{x}$ based on $\mathbf{z}$. The reconstruction loss thus encourages the encoder to increase the signal-to-noise ratio in $\mathbf{z}$ by decreasing $||\boldsymbol \sigma^2_\phi(\mathbf{x}) ||$  and increasing $||\boldsymbol \mu_\phi(\mathbf{x})||$. This effect is countered by the regularization loss, which encourages the encoder to increase $||\boldsymbol \sigma^2_\phi(\mathbf{x}) ||$ and decrease $||\boldsymbol \mu_\phi(\mathbf{x})||$.

In many auto-encoding frameworks with regularization,
whether deterministic (e.g. $L_2$ regularization) or stochastic, 
the weight of the regularization loss is manually tuned until the network achieves a certain desirable behavior. 
On the other hand, in Sections~\ref{sec:prob} and~\ref{sec:information} we will see that in a VAE the one-to-one ratio between the reconstruction loss and regularization loss has a probabilistic and information theoretic meaning.
We will refer to the VAE constructed in this section as a \emph{Typical VAE}
and will continue to refer to Equations~\ref{eq:reconloss} and~\ref{eq:regloss} as an illustration.
However, keep in mind that a Typical VAE is just one of many possible examples of how we can choose to construct $p_\theta(\mathbf{z})$, $p_\theta(\mathbf{x}|\mathbf{z})$ and $q_\phi(\mathbf{z}|\mathbf{x})$.

\section{Probabilistic Interpretation of VAEs}
\label{sec:prob}
\subsection{Bayes's Rule}
\label{sec:bayes}
The VAE was first introduced as an auto-encoder for performing Variational Bayes~\cite{vae,rezende2014stochastic}. The negative ELBO can be derived with a simple application of Bayes' Rule:
\begin{align}
    p_\theta(\mathbf{x}) &= \frac{p_\theta(\mathbf{x}|\mathbf{z})p_\theta(\mathbf{z})}{p_\theta(\mathbf{z}|\mathbf{x})}\\
     -\log p_\theta(\mathbf{x}) &= -\log p_\theta(\mathbf{x}|\mathbf{z}) - \log  p_\theta(\mathbf{z}) + \log p_\theta(\mathbf{z}|\mathbf{x})\\
     &=  -\log p_\theta(\mathbf{x}|\mathbf{z}) - \log  p_\theta(\mathbf{z}) + \log q_\phi(\mathbf{z}|\mathbf{x}) - \log q_\phi(\mathbf{z}|\mathbf{x}) + \log p_\theta(\mathbf{z}|\mathbf{x}) \label{eq:addsubtractq}
\end{align}
We can then take the expectation of both sides over $q_\phi(\mathbf{z}|\mathbf{x})$. Since $p_\theta(\mathbf{x})$ is constant over $\mathbf{z}$, $E_{\mathbf{z}\sim q_\phi(\mathbf{z}|\mathbf{x})} \log p_\theta(\mathbf{x}) = \log p_\theta(\mathbf{x})$. Hence:
\begin{align}
     \begin{split}
         -\log p_\theta(\mathbf{x}) &= E_{\mathbf{z}\sim q_\phi(\mathbf{z}|\mathbf{x})} [-\log p_\theta(\mathbf{x}|\mathbf{z})] +\\
         &\qquad D_{KL}( q_\phi(\mathbf{z}|\mathbf{x}) ||   p_\theta(\mathbf{z})) - D_{KL}(q_\phi(\mathbf{z}|\mathbf{x}) || p_\theta(\mathbf{z}|\mathbf{x}))\\
     \end{split} \label{eq:variationalbayes}
\end{align}
Unfortunately, we cannot efficiently evaluate Equation~\ref{eq:variationalbayes} exactly since  we do not know $p_\theta(\mathbf{z}|\mathbf{x})$.
However, KL-Divergence is non-negative, so we can remove the last term in Equation~\ref{eq:variationalbayes} to obtain an upper bound equal to the negative log likelihood, which yields the negative ELBO:
\begin{align}
     -\log p_\theta(\mathbf{x}) &\leq E_{\mathbf{z}\sim q_\phi(\mathbf{z}|\mathbf{x})} [-\log p_\theta(\mathbf{x}|\mathbf{z})] +
          D_{KL}( q_\phi(\mathbf{z}|\mathbf{x}) ||   p_\theta(\mathbf{z}))
\end{align}
with equality holding if and only if $D_{KL}(q_\phi(\mathbf{z}|\mathbf{x}) ||  p_\theta(\mathbf{z}|\mathbf{x})) = 0$ (i.e. when the encoder is able to perfectly predict $p_\theta(\mathbf{z}|\mathbf{x})$).

\subsection{Importance Sampling}
\label{sec:importance}
Importance-Weighted Auto-Encoders~\cite{burda2015importance} use importance sampling to provide a similar derivation to Section~\ref{sec:bayes} for the negative ELBO by switching the order in which the expectation and logarithm are applied:
\begin{align}
    -\log p_\theta(\mathbf{x}) &=-\log  \int p_\theta(\mathbf{x}|\mathbf{z}) p_\theta(\mathbf{z}) d\mathbf{z}\\
    &= -\log \int \frac{p_\theta(\mathbf{x}|\mathbf{z}) p_\theta(\mathbf{z}) q_\phi(\mathbf{z}|\mathbf{x})}{q_\phi(\mathbf{z}|\mathbf{x})} d\mathbf{z}\\
 &=-\log \mathbb{E}_{\mathbf{z}\sim q_\phi(\mathbf{z}|\mathbf{x})} \frac{p_\theta(\mathbf{x}|\mathbf{z}) p_\theta(\mathbf{z})}{q_\phi(\mathbf{z}|\mathbf{x})} \label{eq:expectationq}
\end{align}
We can then apply Jensen's Inequality to switch the expectation with the logarithm, obtaining an upper bound on the negative log likelihood, which then simplifies to the negative ELBO:
\begin{align}
    -\log p_\theta(\mathbf{x})  &\leq \mathbb{E}_{z\sim q_\phi(\mathbf{z}|\mathbf{x})} -\log \frac{p_\theta(\mathbf{x}|\mathbf{z}) p_\theta(\mathbf{z})}{q_\phi(\mathbf{z}|\mathbf{x})} \label{eq:jensen}\\
    &=\mathbb{E}_{\mathbf{z}\sim q_\phi(\mathbf{z}|\mathbf{x})} [-\log p_\theta(\mathbf{x}|\mathbf{z})]
		 + D_{KL}(q_\phi(\mathbf{z}|\mathbf{x}) || p_\theta(\mathbf{z})) 
\end{align} 

While it is not apparent under this derivation that the tightness of the negative ELBO can be quantified using $D_{KL}(q_\phi(\mathbf{z}|\mathbf{x}) ||  p_\theta(\mathbf{z}|\mathbf{x}))$, we can see that equality holds if $q_\phi(\mathbf{z}|\mathbf{x}) = p_\theta(\mathbf{z}|\mathbf{x})$ since the right-hand-side of Equation~\ref{eq:expectationq} becomes $-\log \mathbb{E}_{\mathbf{z} \sim p_\theta(\mathbf{z}|\mathbf{x})} \frac{p_\theta(\mathbf{x}|\mathbf{z}) p_\theta(\mathbf{z})}{p_\theta(\mathbf{z}|\mathbf{x})} =  - \log \mathbb{E}_{\mathbf{z} \sim p_\theta(\mathbf{z}|\mathbf{x})} p_\theta(\mathbf{x}) = - \log  p_\theta(\mathbf{x}) $.

The key advantage of the importance sampling interpretation is that Equation~\ref{eq:expectationq} gives us a method to approximate the true negative log-likelihood without knowledge of $p_\theta(\mathbf{z}|\mathbf{x})$. In the asymptotic case where we take infinite samples, the approximation can become arbitrarily close. 
However, if the inferred posterior $q_\phi(\mathbf{z}|\mathbf{x})$ deviates too much from the true posterior $p_\theta(\mathbf{z}|\mathbf{x})$, importance sampling may require a prohibitively large number of samples to be accurate.

\subsection{VAEs and Normalizing Flow}
\label{sec:flow}
An alternative maximum likelihood model to VAEs are the more expressive but heavyweight normalizing flow models~\cite{nice}.
In this section, we will look at how a VAE with a Gaussian inferred posterior (but not necessarily diagonal, as in a Typical VAE) could in principle match the performance of normalizing flow models.
However, such a VAE would be computationally no easier to train than a flow model. 

Normalizing flow leverages the change-of-variables formula: let $f$ be an invertible function and $J$ be the Jacobian of $f$. The negative log likelihood of $\mathbf{x}$ can be evaluated using:
\begin{align}
    -\log p(\mathbf{x}) &= -\log p(f^{-1}(\mathbf{x})) - \log |J^{-1}| \label{eq:cov}
\end{align}
where $|J^{-1}|$ is the absolute value of the determinant of $J^{-1}$.
Normalizing flow models learn neural networks that are guaranteed to be invertible and for which $|J^{-1}|$ can be efficiently computed. An invertible mapping $f$ is learned  from a latent space $\mathbf{z}$ (where inference and generation can be easily done on $p_\theta(\mathbf{z})$) to the data space $\mathbf{x}$.
 Equation~\ref{eq:cov} can then be efficiently used as an objective function for maximum likelihood. 

Consider a VAE where $p_\theta(\mathbf{z})$ is an isotropic Gaussian as in a Typical VAE, but the covariance matrix of $q_\phi(\mathbf{z}|\mathbf{x})=\mathcal{N}(\boldsymbol \mu, \Sigma)$ is no longer restricted to be diagonal.
For simplicity, we will assume that the output distribution is also an isotropic Gaussian (i.e. for all $\mathbf{z}$,  $\boldsymbol \sigma^2_\theta(\mathbf{z})=\sigma^2_\theta \mathbf{1}$ where $\sigma^2_\theta$ is a scalar).
We will refer to such a VAE as a $\Sigma$-VAE.
We will now show that for any normalizing flow model that learns an invertible function $f$ from $\mathbf{z}$ to $\mathbf{x}$, a $\Sigma$-VAE can in theory approach a solution where the negative ELBO would be equivalent to Equation~\ref{eq:cov}.

Consider a $\Sigma$-VAE where $\boldsymbol \mu_\theta(\mathbf{z}) = f(\mathbf{z})$, $\boldsymbol \mu_\phi(\mathbf{x}) = f^{-1}(\mathbf{x})$, the covariance matrix is $\Sigma_\phi(\mathbf{x}) = \epsilon^2 J^{-1} {J^{-1}}^T$, and $\sigma^2_\theta = \epsilon^2$ where $\epsilon^2>0$ is a small scalar. Note that since $f$ is invertible, $\mathbf{z}$ must have the same dimensionality as $\mathbf{x}$. Then we can sample a random vector $\mathbf{u}\sim \mathcal{N}(0,\mathbf{I})$ and use the reparameterization trick to sample from $q_\phi(\mathbf{z}|\mathbf{x}) = \mathcal{N}(\boldsymbol \mu_\phi(\mathbf{x}), \Sigma_\phi(\mathbf{x}))$, in which case
\begin{align}
    \mathbf{z} &= f^{-1}(\mathbf{x}) + \epsilon J^{-1} \mathbf{u} 
\end{align}
By definition of the Jacobian, for any vectors $\mathbf{z},\mathbf{v}$, we have:
\begin{align}
    \frac{f(\mathbf{z}+\epsilon \mathbf{v}) - f(\mathbf{z})}{\epsilon} = J\mathbf{v} + \mathbf{r}
\end{align}
for some remainder vector $\mathbf{r}$ where $\lim_{\epsilon \to 0} \mathbf{r} = 0$. This gives us:
\begin{align}
    \boldsymbol \mu_\theta(\mathbf{z}) &= f(f^{-1}(\mathbf{x}) + \epsilon J^{-1} \mathbf{u}) \\
    &= f(f^{-1}(\mathbf{x}))+ \epsilon J J^{-1} \mathbf{u} + \epsilon \mathbf{r}\\
    &= \mathbf{x} + \epsilon \mathbf{u} +  \epsilon \mathbf{r}
\end{align}
Then according to Equation~\ref{eq:reconloss}, our reconstruction loss $L_{rec}$ is given by:
\begin{align}
    L_{rec} &= \mathbb{E}_{\mathbf{u}\sim \mathcal{N}(0,\mathbf{I})} [\frac{n}{2}\log(2 \pi \epsilon^2) + \frac{ ||\mathbf{x} + \epsilon \mathbf{u} +  \epsilon \mathbf{r} - \mathbf{x}||_2^2}{2\epsilon^2}] \\
   & = \frac{1}{2} [n\log(2 \pi ) + n\log \epsilon^2 +  \mathbb{E}_{\mathbf{u}\sim \mathcal{N}(0,\mathbf{I})}[||\mathbf{u}-\mathbf{r}||_2^2]]
\end{align}
The regularization loss is given by:
\begin{align}
    L_{reg} &= D_{KL} ( \mathcal{N}(\boldsymbol \mu_\phi(\mathbf{x}), \Sigma_\phi(\mathbf{x})) || \mathcal{N}(0,\mathbf{I}) )\\
    &= \frac{1}{2}[|| \boldsymbol \mu_\phi(\mathbf{x}) ||_2^2 - \log |\epsilon^2 J^{-1} {J^{-1}}^T| - n + \text{trace}(\epsilon^2 J^{-1} {J^{-1}}^T) ]\\
    &=  \frac{1}{2}[|| f^{-1}(\mathbf{x}) ||_2^2 - \log |J^{-1}|^2 - n \log \epsilon^2 - n + \epsilon^2 \text{trace} ( J^{-1} {J^{-1}}^T)]
\end{align}
Then the total objective function is given by:
\begin{align}
    L &=L_{rec}+L_{reg}\\
    \begin{split}
        &= \frac{1}{2} [ n\log(2 \pi ) + || f^{-1}(\mathbf{x}) ||_2^2] - \log |J^{-1}|^2  + \\
       &\qquad  \mathbb{E}_{\mathbf{u}\sim \mathcal{N}(0,\mathbf{I})}[||\mathbf{u}-\mathbf{r}||_2^2] - n + \epsilon^2 \text{trace} ( J^{-1} {J^{-1}}^T)]\\
    \end{split} 
\end{align}
If we take the limit as $\epsilon \to 0$, then $\mathbf{r}$ approaches $0$, so  $\lim_{\epsilon\to 0} \mathbb{E}_{\mathbf{u}\sim \mathcal{N}(0,\mathbf{I})}[||\mathbf{u}-\mathbf{r}||_2^2] = \mathbb{E}_{\mathbf{u}\sim \mathcal{N}(0,\mathbf{I})}||\mathbf{u}||_2^2 = n$.
Since $\text{trace} ( J^{-1} {J^{-1}}^T)$ is constant with respect to $\epsilon$, $\lim_{\epsilon\to 0} \epsilon^2 \text{trace} ( J^{-1} {J^{-1}}^T) = 0$.
Thus 
\begin{align}
    \lim_{\epsilon\to 0} L &= \frac{1}{2} [ n\log(2 \pi ) + || f^{-1}(\mathbf{x}) ||_2^2 - \log |J^{-1}|^2 ] \\ 
    &= -\log p_\theta(f^{-1}(\mathbf{x})) - \log |J^{-1}| \label{eq:simplifycov}
\end{align}
which is the same as Equation~\ref{eq:cov}. 
Note that the simplification in Equation~\ref{eq:simplifycov} can be done because $p_\theta(\mathbf{z}) = \mathcal{N}(0,\mathbf{I})$.
Since the negative ELBO approaches the true negative log-likelihood as $\epsilon\to 0$, we can also conclude that $\lim_{\epsilon\to 0} D_{KL}( q_\phi(\mathbf{z}||\mathbf{x}) || p_\theta(\mathbf{z}||\mathbf{x})) = 0$.

Hence, a $\Sigma$-VAE could in principle match the modelling capacity of any normalizing flow model. However, in practice there would be a computational bottleneck in computing the determinant of the Jacobian when calculating the regularization loss, meaning that training a $\Sigma$-VAE would be no easier than training a normalizing flow model.

If we force the covariance matrix $\Sigma$ to be a diagonal matrix as in  a Typical VAE, then inference becomes efficient, but log-likelihood can only be approximated as described in this section for functions such that $J^{-1} {J^{-1}}^T$ is diagonal. If we consider the SVD of $J^{-1} = UDV^T$, we see that the covariance matrix is diagonal when $VD^2V^T$ is diagonal.
This can happen when all the eigenvalues of $J$ are equal or when $V$ is a permutation matrix. $V$ is a permutation matrix if each of the dimensions in the latent vector $\mathbf{z}$ influence $\mathbf{x}$ independently.
Thus, though restricting $\Sigma$ to be a diagonal matrix limits the class of functions we can model, it may also naturally encourage disentanglement. 
The relationship between VAEs and disentanglement remains an active area of research~\cite{betavae,chen2018isolating}.
We also highlight that regardless of whether $\Sigma$ is diagonal or not, the Gaussian modelling assumptions of our model do not restrict the modelling capacity of a VAE to only Gaussian-like data.

\section{Information Theoretic Interpretation of VAEs}
\label{sec:information}

\subsection{Lossless Compression}
\label{sec:lossless}

Maximum likelihood can also be viewed as a lossless compression problem.
The negative log likelihood $- \mathbb{E}_{\mathbf{x} \sim p_{gt}(\mathbf{x})} \log p_\theta(\mathbf{x}) $ is the optimal expected number of bits needed to describe a sample from $p_{gt}(\mathbf{x})$ based on $p_\theta(\mathbf{x})$.
VLAE~\cite{vlae} shows that the negative ELBO $-\mathcal{L}(\mathbf{x})$ is an upper bound on this number by constructing a code to describe $p_{gt}(\mathbf{x})$ that on average uses $-\mathcal{L}(\mathbf{x})$ bits.

Suppose a sender and receiver have access to our VAE, and the sender wishes to send a vector $\mathbf{x} \sim p_{gt}(\mathbf{x})$ to the receiver. The sender could first sample $\mathbf{z}\sim q_\phi(\mathbf{z}|\mathbf{x})$. Sending $\mathbf{z}$ will cost $-\log p_\theta(\mathbf{z})$ bits. The receiver can then use the VAE to decode $p_\theta(\mathbf{x}|\mathbf{z})$. The sender can then spend another $-\log p_\theta(\mathbf{x}|\mathbf{z})$ bits to send an additional code (e.g. the error-correcting code of a reconstruction) that could then describe $\mathbf{x}$ exactly. The sender on average spends $-\log p_\theta(\mathbf{z}) -  \log p_\theta(\mathbf{x}|\mathbf{z})$ bits to describe $\mathbf{x}$.
However, since the receiver now knows $\mathbf{x}$, it can then use the VAE to know $q_\phi(\mathbf{z}|\mathbf{x})$, from which it can decode a secondary message. For example, if the VAE is reparameterized as in Equation~\ref{eq:reparameterize}, then the receiver now also knows the value of $\mathbf{u}$, which contains $-\log q_\phi(\mathbf{z}|\mathbf{x})$ bits of information.
The expected cost $C$ used to describe $\mathbf{x}$ with this coding scheme is thus given by:
\begin{align}
    C = \mathbb{E}_{\mathbf{z}\sim q_\phi(\mathbf{z}|\mathbf{x})}[-\log p_\theta(\mathbf{z}) -  \log p_\theta(\mathbf{x}|\mathbf{z}) +  \log q_\phi(\mathbf{z}|\mathbf{x})]
\end{align}
which is equal to the negative ELBO.

Based on this interpretation we can intuitively understand how the reconstruction loss and regularization loss interact with each other in a Typical VAE. The reconstruction loss essentially copies information from the input space to the latent space, and the regularization loss compresses that information.
Even if information in the input space is incompressible (i.e. because it is pure noise), copying it into to the latent space will not hurt the negative log likelihood (although it will not help either).
Thus, in an ideal  optimization landscape, during training we would generally expect the reconstruction loss to decrease and the regularization loss to increase as more information gets stored in the latent space. Although VAEs are sometimes stereotypically associated with blurry reconstructions and heavy stochastic noise in the latent space, in reality as training progresses the VAE should exhibit essentially perfect reconstructions and increasingly deterministic behavior as training progresses.

\subsection{Continuous versus Discrete Data}
\label{sec:discrete}
So far our discussion has been limited to continuous data.
However, in practice the data we work with is often discrete or quantized, so our discussion is incomplete without considering to what degree of accuracy we wish to describe $\mathbf{x}$.
For example, we typically wish to describe $RGB$ images to $8$-bit accuracy.
Hence, assuming that our data is normalized to lie in the range $[0,1]$, the cost $L_{rec}$ to describe the reconstruction error for an image using the coding scheme in Section~\ref{sec:lossless} would more accurately be written as:
\begin{align}
    L_{rec}(\mathbf{x},\mathbf{z}) &= - \log \int_{\mathbf{u} \in [0,\frac{1}{256})^n} p_\theta(\mathbf{x}+\mathbf{u} | \mathbf{z}) d\mathbf{u}  \label{eq:discrete}
\end{align}
In Section~\ref{sec:lossless},  we discussed that in a nice optimization landscape, we expect the reconstruction loss to decrease and the regularization loss to increase during training.
However, for discrete data, this can only occur until $L_{rec}$ reaches $0$, at which point the reconstruction loss can no longer decrease and the regularization loss will begin to decrease as compression occurs.

Since Equation~\ref{eq:discrete} requires taking an integral,
when working with discrete data it may be beneficial to model $-\log p_\theta(\mathbf{x}|\mathbf{z})$ using a distribution for which taking the cumulative distribution function (CDF) is efficient and differentiable.
Thus instead of modelling $-\log p_\theta(\mathbf{x}|\mathbf{z})$ with an isotropic Gaussian as in a Typical VAE, VAE-IAF~\cite{kingma2016improved} instead uses the similarly bell-shaped isotropic logistic distribution, for which the CDF is given by the sigmoid function.
 
 The units for evaluating a maximum likelihood model for image data is bits per dim (bits/dim), which is the number of bits such a model would need to losslessly describe each pixel of the image to $8$-bit accuracy.
 Any model with a negative log-likelihood of more than $8$ bits/dim is worse than useless as describing the raw pixel values only requires $8$ bits/dim.
 Current state-of-the-art maximum likelihood models require a little less than $3$ bits/dim~\cite{child2019generating}.
 
 \subsection{Transmission Across a Noisy Gaussian Channel}
\label{sec:transmission}
 
Once $L_{rec}$ reaches $0$, we can draw an analogy between a VAE and the classical problem of transmission across a memoryless noisy Gaussian channel.
In such a problem, a sender wishes to reliably  describe $\mathbf{x}$ by transmitting a continuous number $\hat{z}$ across a noisy Gaussian channel $d$ times. 
However, due to power constraints, the sender can only send a strong enough signal such that $\mathbb{E} [\hat{z}^2] = P$.
Moreover, every time the sender sends a signal, the channel adds Gaussian noise $y$ such that $\mathbb{E} [y^2] = N$.
Hence the receiver receives $d$ transmissions of $z = \hat{z} + y$. If $P+N=1$, then $\mathbb{E} [z^2] = 1$.
The capacity $R$ of a Gaussian channel with noise $N$ and power $P=1-N$ is given by~\cite{cover2012elements}:
\begin{align}
    R &=-\frac{1}{2}\log(N)
\end{align}
If the channel has capacity $R$, then $dR$ bits of information can reliably be transmitted across the channel with arbitrarily low probability of error as $d$ increases.  Transmission across a noisy channel can thus be viewed as the dual problem of compression, since the ability to describe reliably $\mathbf{x}$ using $d$ transmissions of a channel with capacity $R$ indicates an ability to compress $\mathbf{x}$ into a representation of at most $dR$ bits.

In an analogous VAE, $p_\theta(\mathbf{z}) = \mathcal{N}(0,\mathbf{I})$, $q_\phi(\mathbf{z}|\mathbf{x}) = \mathcal{N}(\boldsymbol \mu_\phi(\mathbf{x}), N_\phi(\mathbf{x}) \mathbf{I})$ where $N_\phi(\mathbf{x})>0$ is a scalar, $\mathbf{x} \in \mathbb{R}^n, \mathbf{z} \in \mathbb{R}^d$.
 $P_\phi(\mathbf{x})= \frac{||\boldsymbol \mu_\phi(\mathbf{x})||_2^2}{d} $ corresponds to the power, each dimension of $\boldsymbol \mu_\phi(\mathbf{x})$ corresponds to a transmission $\hat{z}$ across the noisy channel, $N_\phi(\mathbf{x})$ corresponds to the noise, and each dimension of $\mathbf{z}$ corresponds to a received transmission $z$. If we set $N_\phi(\mathbf{x})=1-P_\phi(\mathbf{x})$, then according to Equation~\ref{eq:regloss}, the expected regularization loss $L_{reg}$ is given by:
\begin{align}
    L_{reg}(\mathbf{x}) & = 
    \frac{1}{2} ||\boldsymbol \mu_\phi(\mathbf{x})||_2^2 + \frac{1}{2} [ d N_\phi(\mathbf{x}) - d - d \log N_\phi(\mathbf{x})]\\
    & =  \frac{1}{2} dP_\phi(\mathbf{x}) + \frac{1}{2} [ dN_\phi(\mathbf{x})  - d - d \log N_\phi(\mathbf{x})] \\
    & =-\frac{1}{2}d\log(N_\phi(\mathbf{x}))
\end{align}
which is equal to $dR$ in the transmission problem. 

We now give an intuitive geometric explanation for why a channel with power $P$ and noise $N$ (where $P+N=1$) has capacity $R=-\frac{1}{2}\log(N)$. A formal proof is given in~\cite{cover2012elements}.
By the law of large numbers, as $d$ increases,  $||\mathbf{z}||=1$ for almost all $\mathbf{z}\sim \mathcal{N}(0,\mathbf{I})$, so $\mathbf{z}$ essentially forms a uniform distribution on the surface of the unit hyper-sphere. Let $V$ be the volume of the unit hyper-sphere.
Similarly, $\mathbf{z}$ will lie uniformly on the surface of the hyper-sphere with radius $\sqrt{N}$ and center $\boldsymbol \mu$ for almost all $\mathbf{z}\sim \mathcal{N}(\boldsymbol \mu, N \mathbf{I})$. The volume of a hyper-sphere with radius $\sqrt{N}$ is given by $V_N = V N^{\frac{d}{2}}$.
We can thus expect to fit $\frac{V}{V_N} = N^{-\frac{d}{2}} = 2^{dR}$ unique non-overlapping hyper-spheres of radius $\sqrt{N}$ into the unit hyper-sphere.

We can thus create a transmission scheme as follows.
A codebook is created that maps every point $\mathbf{x}$ to a hyper-sphere with volume proportional to $p_{gt}(\mathbf{x})$ and revealed to both the sender and receiver.
Each hyper-sphere has center $\boldsymbol \mu$ and radius $\sqrt{N}$ such that $N + \frac{||\boldsymbol \mu||_2^2}{d} =1$.
During transmission, the sender maps each input $\mathbf{x}$ to the center $\boldsymbol \mu$ of its corresponding hyper-sphere, which is transmitted across the channel.
The receiver then uses the codebook to map each received point $\mathbf{z}$ to the center $\boldsymbol \mu$ of the hyper-sphere that  $\mathbf{z}$ belongs to, which can then be mapped to the input image $\mathbf{x}$.

Since there are finitely many values $\mathbf{x}$ can take when it is discrete, constructing such a codebook is possible in theory. However, in practice, for a $32\times 32$ resolution RGB image, there would be $256^{3072}$ possible values of $\mathbf{x}$, so calculating such a codebook by brute force would be prohibitively expensive.
Optimizing a neural network to learn such a coding scheme would also be difficult since mapping a point to the center of the hyper-sphere that it belongs to would be a non-diffierential operation.
VQ-VAE~\cite{oord2017neural} takes a step in this direction by using the straight-through gradient estimator.

One counter-intuitive result of information theory is that the rate of a memoryless noise channel is optimal, meaning that we could do no better if we were able to send each transmission $\hat{z}_i$ with knowledge of what the receiver received for $z_j$ for all all $j<i$. In other words, when $d$ is large enough, the modelling capacity of a VAE is not inhibited by the assumption that the inferred posterior has diagonal covariance, and in theory a Typical VAE should be able to model any distribution to arbitrary accuracy. 
Indeed, in the above transmission scheme $q_\phi(\mathbf{z}|\mathbf{x}) =p_\theta(\mathbf{z}|\mathbf{x})$, so the negative ELBO would be an exact estimate of the negative log-likelihood.
This is despite making no assumptions about whether $p_{gt}(\mathbf{x})$ is Gaussian or that the latent dimensions of $\mathbf{z}$ affect $\mathbf{x}$ independently.
This is consistent with the results of Section~\ref{sec:flow} since the mapping from $\mathbf{z}$ to $\mathbf{x}$ is not invertible.
However, though VAEs may have the capacity to model a distribution $p_\theta(\mathbf{x})$, in practice they are usually unable to learn such a solution due to computational constraints and a difficult optimization landscape.


\section{Misconceptions About VAEs}
\label{sec:misconceptions}
In this section, we will look at misconceptions about VAEs that I have encountered in teaching materials and among other researchers in computer vision and machine learning. Not all readers may hold these miconceptions, so those who can correctly answer all of the following questions can skip this section:
\begin{itemize}
    \item \textbf{Q:} Can VAEs be trained using mean-squared error as the reconstruction loss? \textbf{A:} No.
    \item \textbf{Q:} Should VAEs have blurry or sharp reconstructions? \textbf{A:} Reconstructions should be essentially perfect.
    \item \textbf{Q:} Are VAEs only able to model data that is highly Gaussian? \textbf{A:} No.
    \item \textbf{Q:} Suppose $p_{gt}(\mathbf{x})$ is an isotropic Gaussian. How many dimensions does the latent space of a Typical VAE  need to be in order to model $p_{gt}$? \textbf{A:} $0$. $\mathbf{z}$ can be ignored and need not exist.
    \item \textbf{Q:} Suppose $p_{gt}(\mathbf{x})$ where $\mathbf{x} \in \mathbb{R}^n$ consists of $k>1$ mixtures of isotropic Gaussians. How many dimensions does the latent space of a Typical VAE need to be in order to model $p_{gt}$? \textbf{A:} $n$. The number of mixtures $k$ is irrelevant.
\end{itemize}

\subsection{VAEs Cannot Be Trained With The Mean-Squared Error Loss}
\label{sec:mse}

Minimizing the Mean-Squared Error ($MSE$) in lieu of $-\log p_\theta(\mathbf{x}|\mathbf{z})$ is a common mistake I have encountered in casual conversation, a  peer review, and at least one prominent blog/code tutorial on VAEs.
In fact, using $MSE$ as the reconstruction loss is extremely problematic for maximum likelihood and results in the VAE being highly over-regularized. 
The $MSE$ objective can be defined as:
\begin{align}
    MSE = \frac{1}{n}\mathbb{E}_{\mathbf{x}\sim p_{gt}(\mathbf{x}), \mathbf{z} \sim q_\phi(\mathbf{z}|\mathbf{x})} ||\mathbf{x} - \boldsymbol \mu_\theta(\mathbf{z})||_2^2
\end{align}
In a simplified Typical VAE where $\boldsymbol \sigma^2_\theta(\mathbf{z}) = \sigma^2_\theta \mathbf{1}$ as described in Section~\ref{sec:flow}, the reconstruction loss is given by:
\begin{align}
    L_{rec} &=  \frac{1}{n}\mathbb{E}_{\mathbf{x}\sim p_{gt}(\mathbf{x}), \mathbf{z} \sim q_\phi(\mathbf{z}|\mathbf{x})}[-\log p_\theta(\mathbf{x}|\mathbf{z})]\\
    &= \mathbb{E}_{\mathbf{x}\sim p_{gt}(\mathbf{x}), \mathbf{z} \sim q_\phi(\mathbf{z}|\mathbf{x})} [ \frac{1}{2}[\log(2\pi\sigma_\theta^2)+\frac{||\mathbf{x}-\boldsymbol \mu_\theta(\mathbf{z})||_2^2}{\sigma_\theta^2}] ]\\
    &=\frac{1}{2}[\log(2\pi\sigma_\theta^2) + \frac{1}{\sigma_\theta^2} MSE] \label{eq:lrecmse}
\end{align}
Note that we have normalized $MSE$ and $-\log p_\theta(\mathbf{x}|\mathbf{z})$ by $\frac{1}{n}$ for simplicity, which is common practice (see Section~\ref{sec:discrete}).

We see that we obtain the $MSE$ objective by assuming $\sigma^2_\theta=\frac{1}{2}$ throughout training and ignoring the first term in Equation~\ref{eq:lrecmse}, which would now be constant.
Such an assumption would obviously be sub-optimal from a maximum likelihood standpoint; in fact we can analytically see that, given $MSE$, Equation~\ref{eq:lrecmse} is minimized when $\sigma^2_\theta = MSE$, in which case $L_{rec}$ simplifies to:
\begin{align}
    L_{rec} = \frac{1}{2}[\log(2 \pi MSE)+1] \label{eq:logmse}
\end{align}
Hence, minimizing negative log-likelihood introduces a logarithm operation in front of the $MSE$ objective.
This should fall in line with our intuition from a lossless compression standpoint, since if the average value of the reconstruction error $|\mathbf{x} - \boldsymbol \mu_\theta(\mathbf{z})|$ were cut in half, which would cut $MSE$ by three-quarters, we would expect to require one fewer bit to describe the reconstruction error.

The central practical difference between using $MSE$ and $\log MSE$ as a reconstruction loss for a VAE is that as the magnitude of $MSE$ decreases by a factor of $c^2$, the gradient of $c^2 MSE$ decreases by a factor of $c$ while the gradient of $\log c^2 MSE = \log c^2 + \log MSE$ remains constant. This results in several problems for $MSE$.

First, since we expect $MSE$ to decrease during training, the gradient of the reconstruction loss relative to the regularization loss becomes weaker and weaker---often by several orders of magnitude---as training progresses. We expect the model to thus be much more highly regularized at the end of training compared to the beginning.

Second, the scale of the input data essentially becomes a hyper-parameter that controls how much we wish to balance the initial weight of the reconstruction loss compared to the regularization loss. For example, we can make $MSE < \epsilon^2$ for arbitrarily small $\epsilon>0$ by simply normalizing all our input data to lie in the range $[0,\epsilon]$. Thus, training a VAE with $MSE$ as the reconstruction loss could more accurately be described as ``an auto-encoder with stochastic regularization that generates useful or beautiful samples".

Using $MSE$ to assess the reconstruction capabilities of a VAE thus becomes quite meaningless. For example, one top tier conference paper incorrectly declared that their VAE had essentially memorized a natural image dataset when they had achieved a $MSE$ of roughly $0.01$, which is an ostensibly low number.
However, since the data was normalized to lie in the range $[0,1]$, a $MSE$ of $0.01$ meant that reconstructions were off by an average of roughly 25 pixel values if normalized back to $0-255$ range. From a lossless compression standpoint as discussed in Section~\ref{sec:discrete}, if we calculated $L_{rec}+8$ using Equation~\ref{eq:logmse}, their model would cost an average of around $7.4$ bits just to describe the reconstruction error of each pixel to $8$-bit accuracy; for context, state-of-the-art VAEs with a diagonal Gaussian posterior can achieve a total negative log-likelihood (including both the reconstruction and the latent cost) of around $3.5$ bits per pixel.

The above problems all point to VAEs trained with $MSE$ being extremely over-regularized during training when the input data is scaled to lie in the range $[0,1]$.
The over-regularization due to incorrectly using $MSE$ instead of $\log MSE$ as a reconstruction loss is also a source of the stereotype that VAEs have ``blurry" reconstructions when in fact a properly trained VAE should have nearly perfect reconstructions as discussed in Section~\ref{sec:lossless}.
Over-regularization can be further amplified when the architecture of the VAE is intentionally restricted so that the latent space is low-dimensional and thus inexpressive (e.g. an hour-glass shaped architecture).

On the other hand, a potential benefit of over-regularization is that the latent space only stores the most highly compressible information, which tend to be global structures. Hence, an over-regularized VAE can also be thought of as a \emph{lossy} compressor. This may be beneficial for related downstream tasks like representation learning or disentanglement. Additionally, the required $MSE$ necessary to achieve state-of-the-art maximum likelihood results is often significantly lower than the threshold of human perception. If our goal is to sample perceptually pleasing images or perform lossy compression, then we may want to over-regularize our model to tolerate higher reconstruction errors (leading to sub-optimal likelihood) at the benefit of better compression and sampleability. A more thorough discussion and an algorithm to control these trade-offs associated with over-regularization can be found in~\cite{rezende2018taming}.

\subsection{The Latent Vector Is Not A Parameter}
\label{sec:latentparameter}
The encoder in a Typical VAE \emph{learns} the parameters $\phi$ of a neural network so that given an input $\mathbf{x}$, it can \emph{predict} the parameters $\boldsymbol \mu_\phi(\mathbf{x}), \boldsymbol \sigma^2_\phi(\mathbf{x})$ of a Gaussian distribution from which $\mathbf{z}$ is sampled.  
The latent vector $\mathbf{z}$ is \emph{not} a parameter that describes the distribution, but rather a \emph{variable} that describes an individual data point $\mathbf{x}$.
Another misconception about VAEs that I have encountered in casual conversation, a peer review, and at least one university graduate course is to think of the parameters $\boldsymbol \mu_\phi(\mathbf{x}), \boldsymbol \sigma^2_\phi(\mathbf{x})$ as \emph{learned} rather than predicted and/or to in turn think of the latent vector $\mathbf{z}$ as a parameter rather than a variable.

To illustrate this, let us consider a toy case of a Typical VAE modelling $p_{gt}(\mathbf{x}) = \mathcal{N}(\boldsymbol \mu_{gt}, \sigma^2_{gt} \mathbf{I})$ where $\mathbf{x}, \mathbf{z} \in\mathbb{R}^n$.
Those who view $\mathbf{z}$ as a parameter may incorrectly guess that the global optimum to ``learn the distribution" involves setting $q_\phi(\mathbf{z}|\mathbf{x}) = \mathcal{N}(\boldsymbol \mu_{gt}, \sigma^2_{gt} \mathbf{I}) = p_{gt}(\mathbf{x})$.
This would clearly be sub-optimal as $q_\phi(\mathbf{z}|\mathbf{x})$ contains no information about $\mathbf{x}$ while incurring a non-zero regularization loss (unless the prior is constructed so that $p_\theta(\mathbf{z}) = p_{gt}(\mathbf{x})$).

Let us consider two valid solutions. One obvious global optimum would be for the VAE to ignore the latent space, so $q_\phi(\mathbf{z}|\mathbf{x}) = p_\theta(\mathbf{z})$ and $p_\theta(\mathbf{x}|\mathbf{z}) = p_\theta(\mathbf{x}) = p_{gt}(\mathbf{x}) = \mathcal{N}(\boldsymbol \mu_{gt}, \sigma^2_{gt} \mathbf{I})$.
Another solution would be for the VAE to learn to essentially act as a deterministic regular auto-encoder, so $q_\phi(\mathbf{z}|\mathbf{x}) = \mathcal{N}(\frac{\mathbf{x} - \boldsymbol \mu_{gt}}{\sigma_{gt}}, \epsilon^2 \mathbf{I})$ and $p_\theta(\mathbf{x}|\mathbf{z}) = \mathcal{N}(\sigma_{gt}\mathbf{z} +\boldsymbol \mu_{gt}, \sigma^2_{gt} \epsilon^2 \mathbf{I})$. It is easy to verify that this solution is a global minimum that achieves the same negative log-likelihood as the first solution as $\epsilon \to 0$.

In both solutions, knowledge about the distribution (i.e. knowledge about $\boldsymbol \mu_{gt}$ and $\sigma^2_{gt}$) are embedded in the parameters $\theta$ and (in the second solution) $\phi$. In the first solution $\mathbf{z}$ is completely uninformative and would still be a valid solution if $\mathbf{z}$ did not exist. 
In the second solution, $\mathbf{z}$ describes the data point $\mathbf{x}$ exactly but contains no information of the distribution or its parameters. As  $\mathbf{z}$ describes a data point, even if $p_{gt}$ were a $k$-mixture of Gaussians, $\mathbf{z}$ would still only need to be $n$-dimensional and not $kn$-dimensional (which would be the case if $\mathbf{z}$ were estimating parameters).

A consequence of incorrectly viewing $\mathbf{z}$ as a parameter is that one will under-estimate the expressivity of a VAE.
Since the information about $p_{gt}$ is all embedded in $\phi$ and $\theta$, which can be arbitrarily powerfully parameterized, VAEs can model complicated non-Gaussian distributions as discussed in Sections~\ref{sec:flow} and~\ref{sec:transmission} and empirically shown in Section~\ref{sec:code}. However, if one incorrectly believes that the purpose of the latent vector $\mathbf{z}$ is to estimate the parameters of a distribution, then one may expect that a VAE can only model data similar to the isotropic Gaussian prior $p_\theta(\mathbf{z})$.

\section{Example: Toy 2D Data}
\label{sec:code}

\begin{figure}[t]
    \centering
    \begin{subfigure}[b]{0.48\textwidth}
        \centering
                \includegraphics[width=\linewidth]{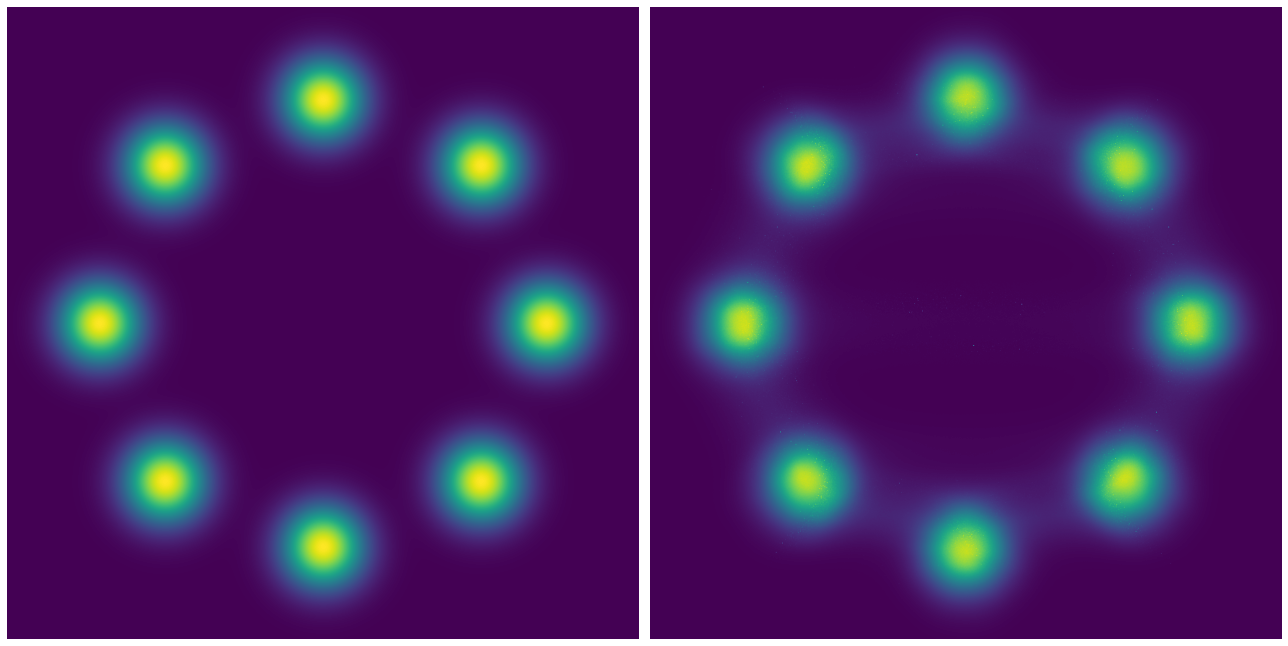}
                \includegraphics[width=\linewidth]{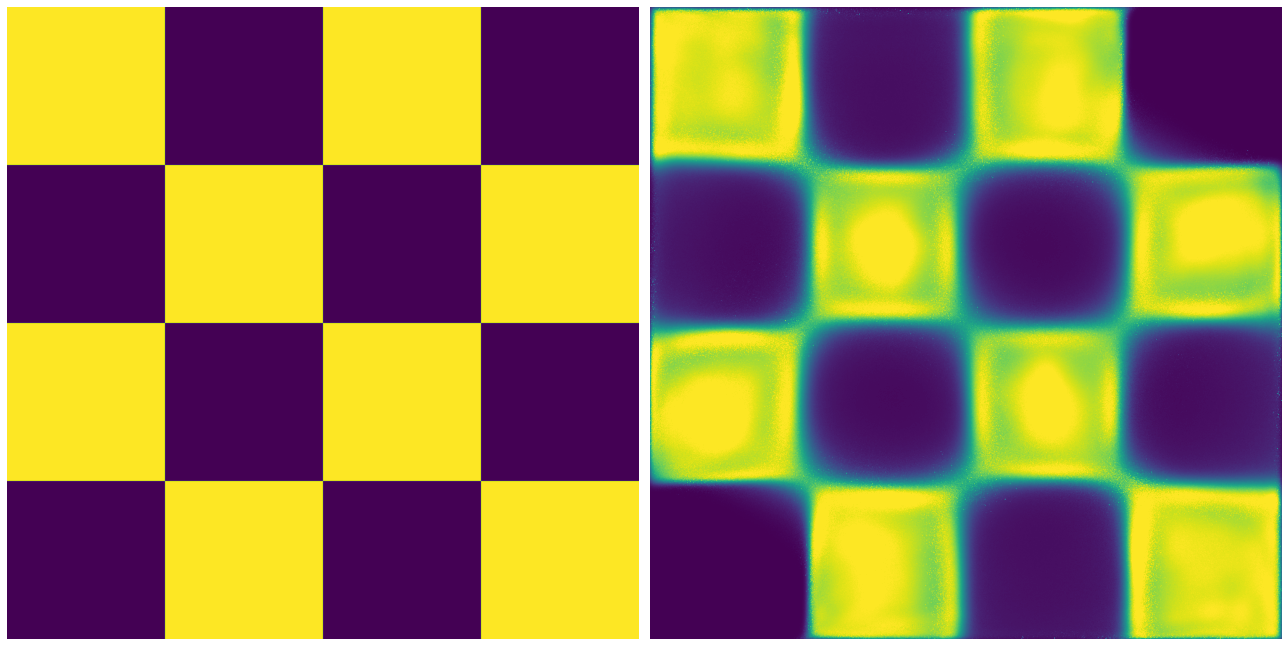}
                \includegraphics[width=\linewidth]{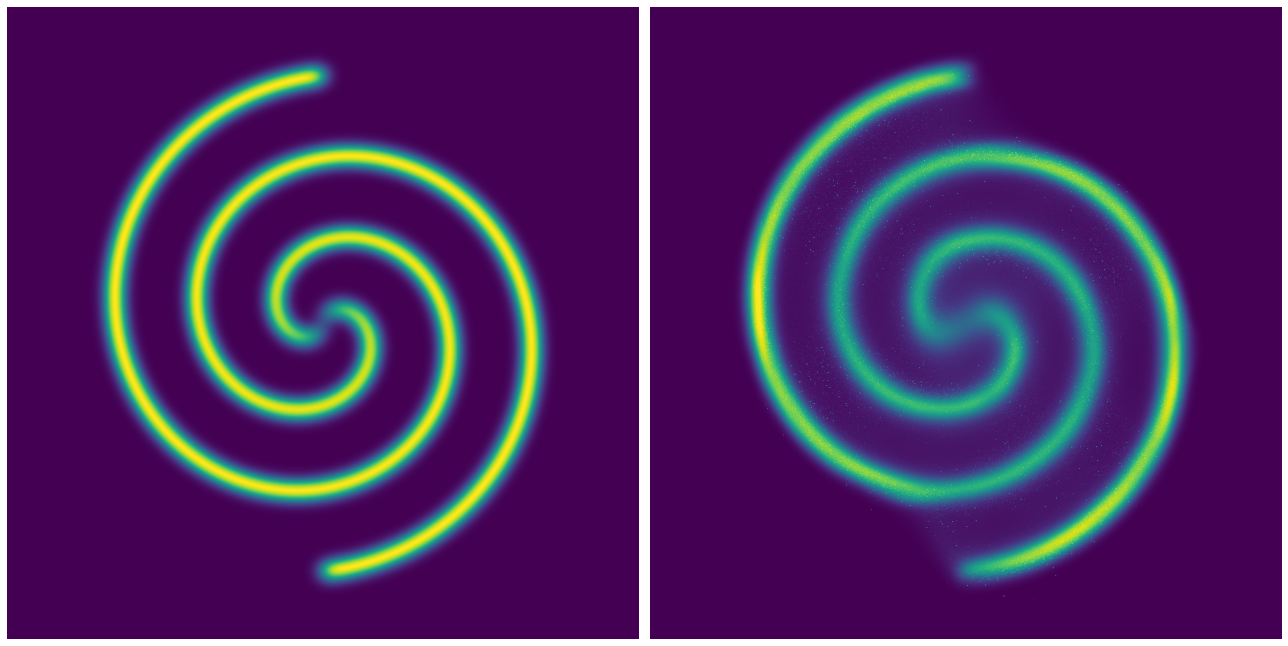}
        \caption{The ground truth probability of each pixel (left) and estimated probability from the VAE (right).} \label{fig:density}
    \end{subfigure}
        \begin{subfigure}[b]{0.48\textwidth}
        \centering
                \includegraphics[width=\linewidth]{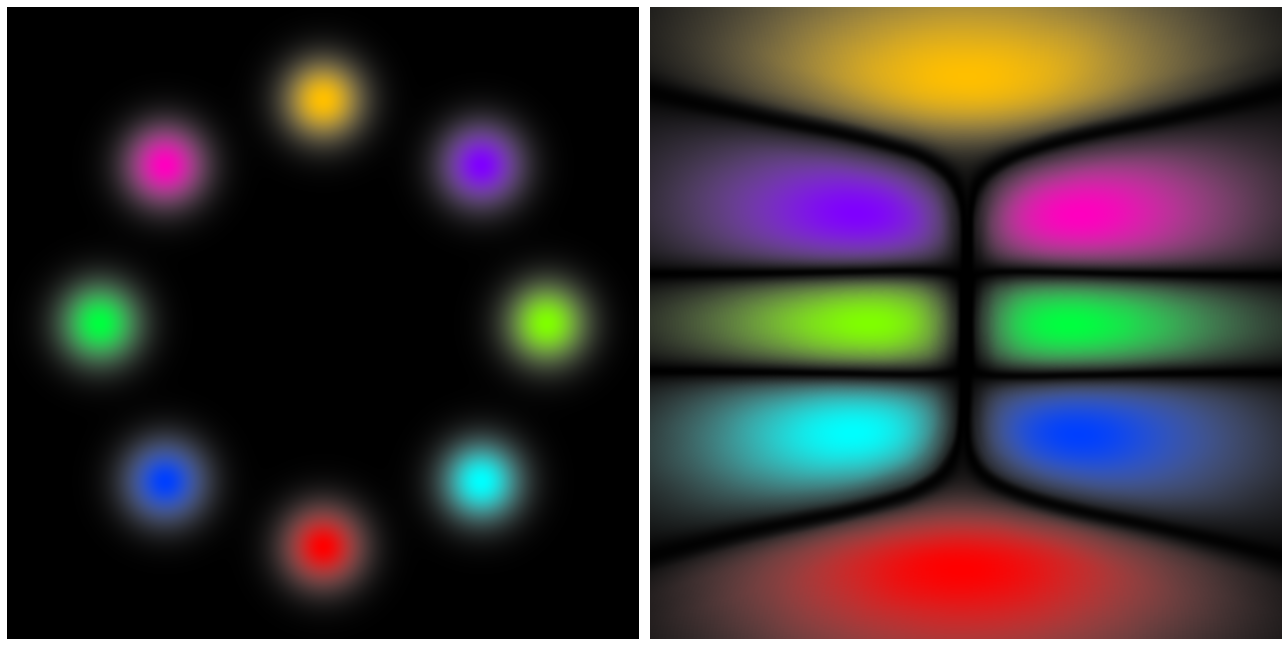}
                \includegraphics[width=\linewidth]{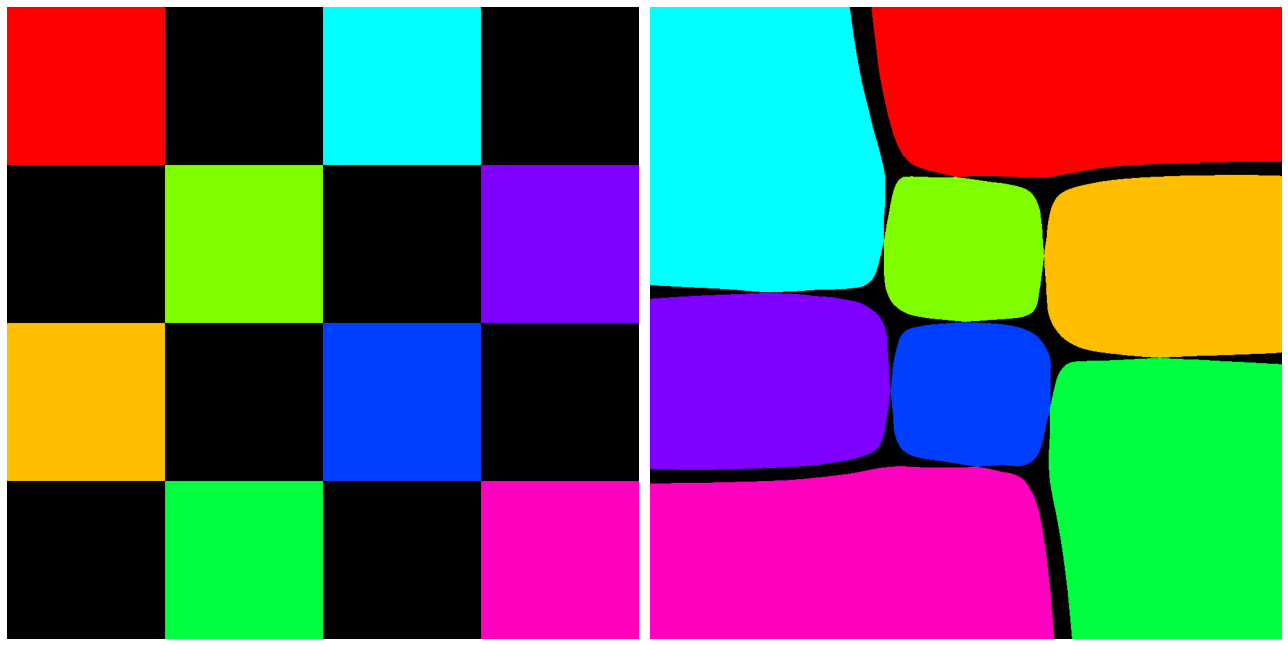}
                \includegraphics[width=\linewidth]{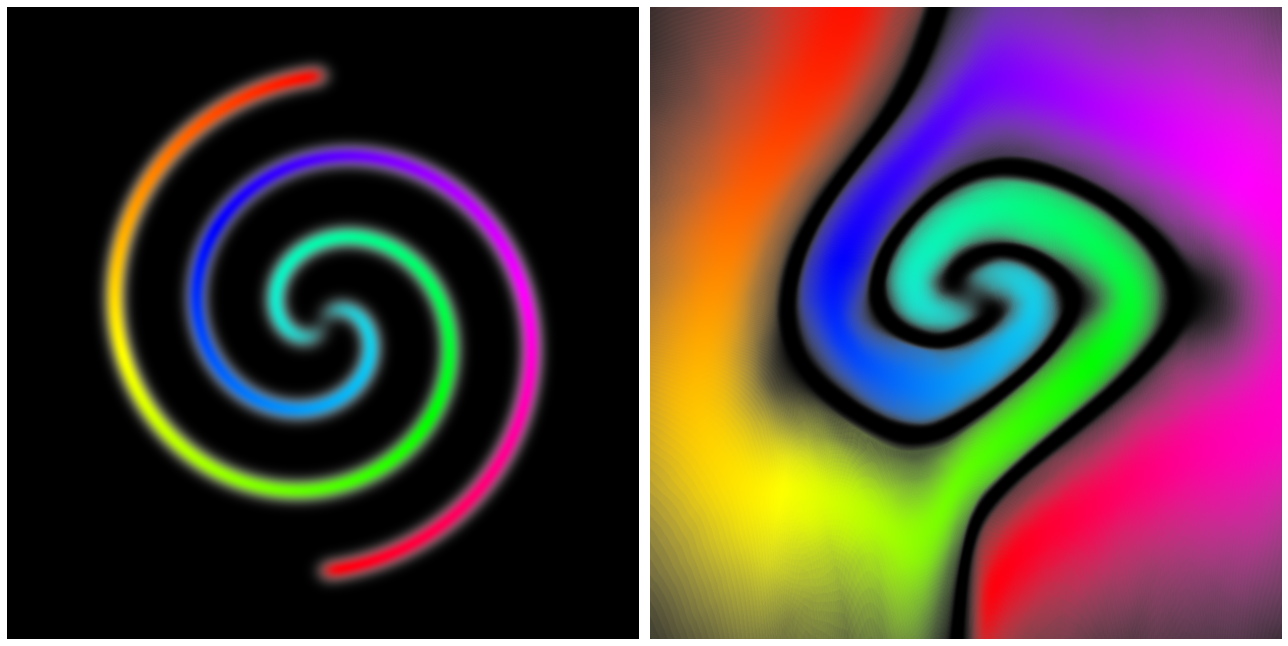}
        \caption{Color correspondence between the ground truth distribution (left) and the latent space (right).}
        \label{fig:latent}
    \end{subfigure}
    \caption{We perform density estimation and visualize the latent space for three 2D distributions. Yellow indicates high probability density and purple indicates low density.}
    \label{fig:typical}
\end{figure}

\begin{figure}[t]
    \centering
    \begin{subfigure}[b]{0.48\textwidth}
        \centering
                \includegraphics[width=\linewidth]{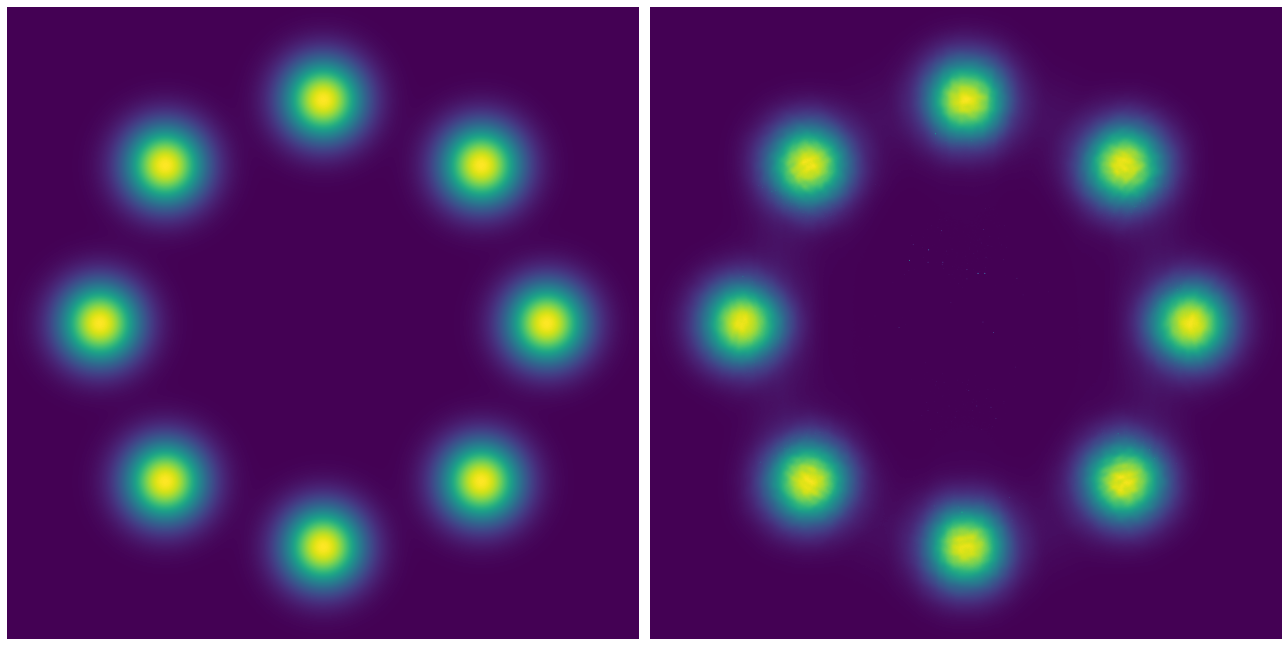}
                \includegraphics[width=\linewidth]{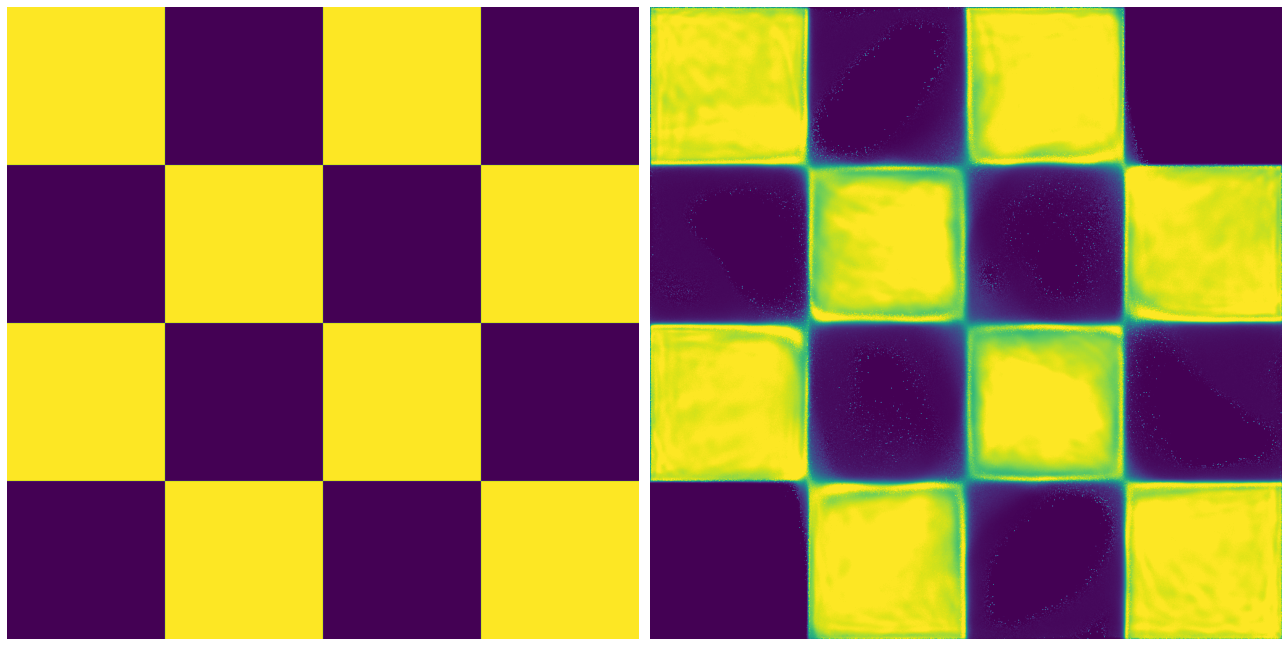}
                \includegraphics[width=\linewidth]{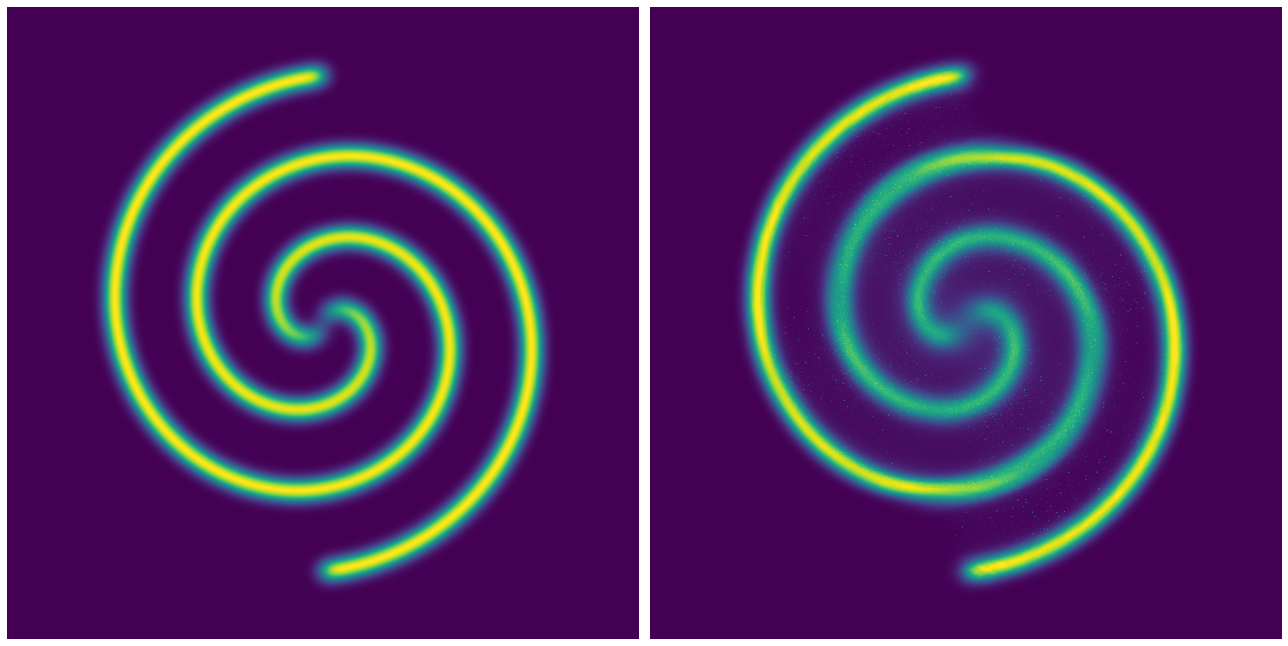}
        \caption{The ground truth probability of each pixel (left) and estimated probability from the VAE (right).} 
    \end{subfigure}
        \begin{subfigure}[b]{0.48\textwidth}
        \centering
                \includegraphics[width=\linewidth]{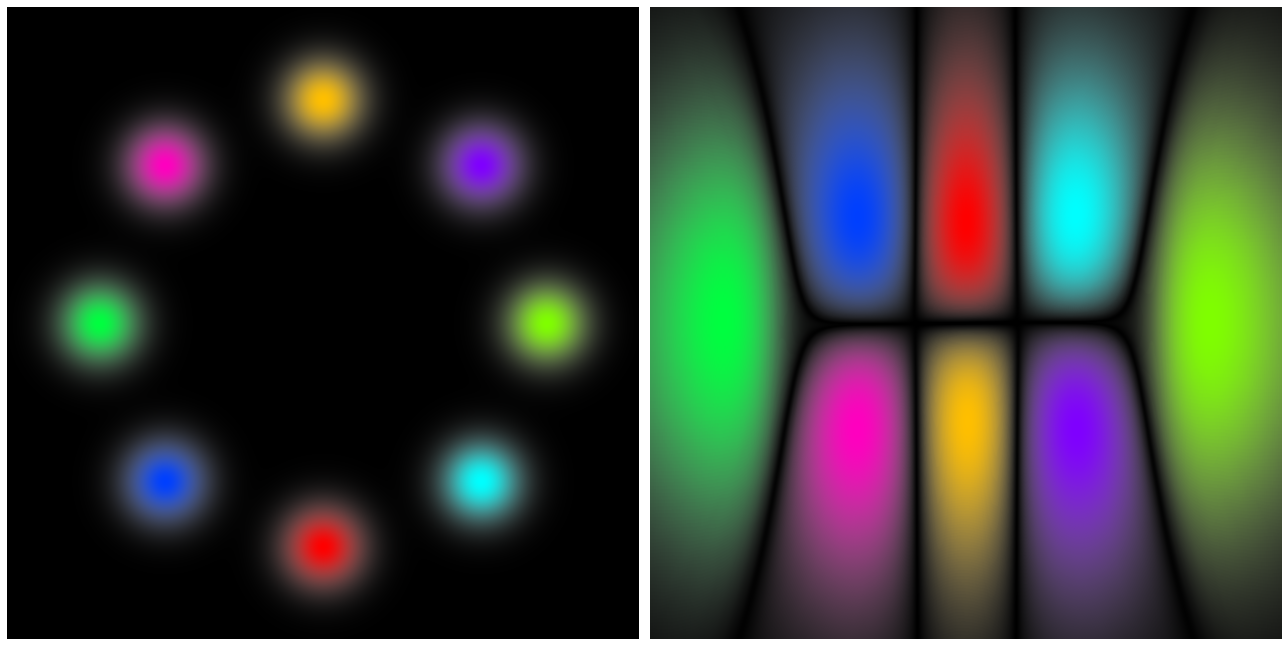}
                \includegraphics[width=\linewidth]{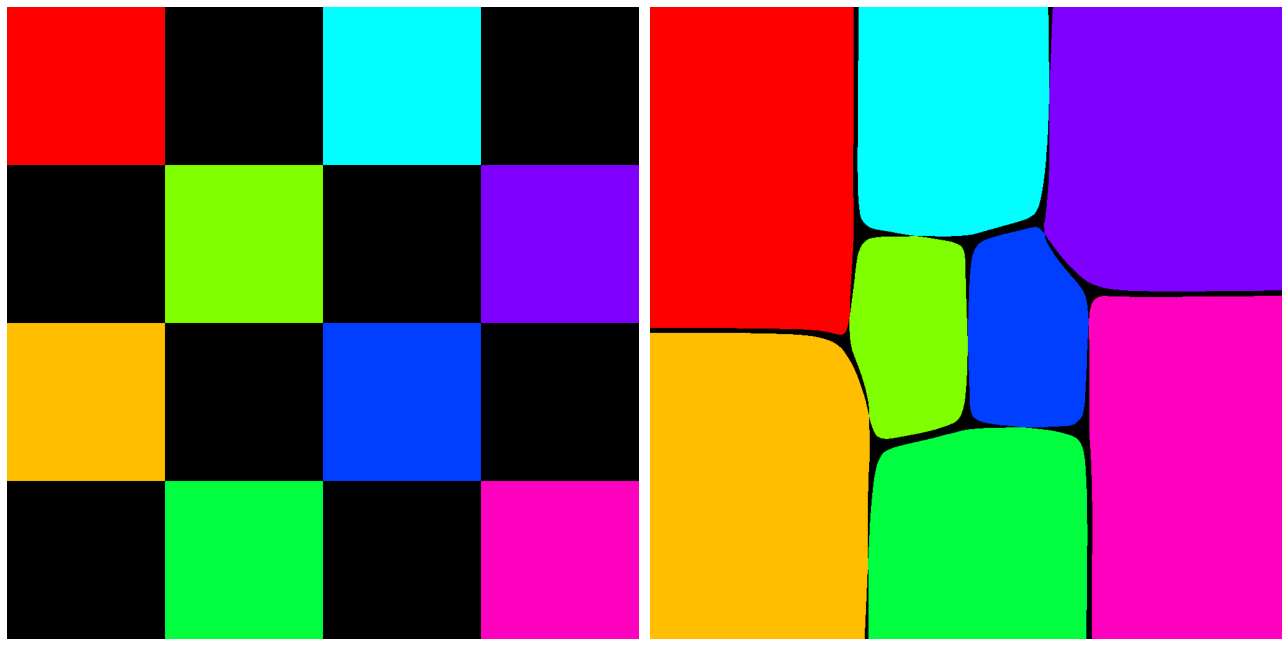}
                \includegraphics[width=\linewidth]{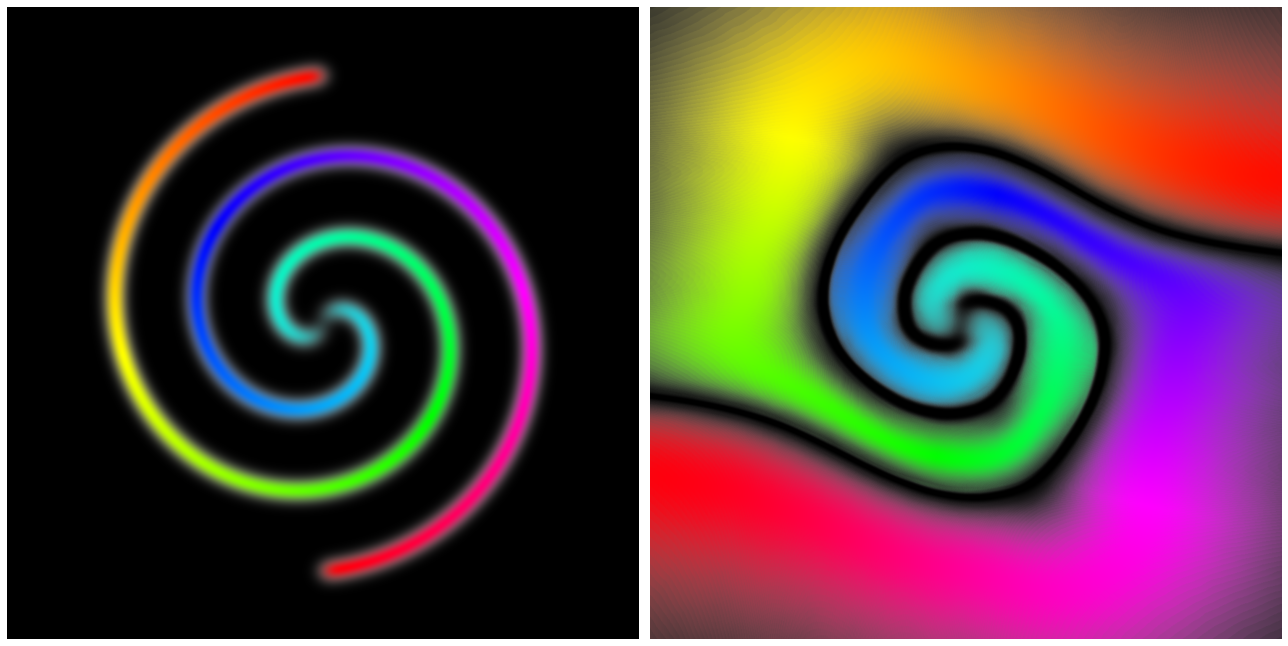}
        \caption{Color correspondence between the ground truth distribution (left) and the latent space (right).}
    \end{subfigure}
    \caption{Density Estimation and Latent Space Visualization for VAEs trained on a large batch size of 40000.}
    \label{fig:big}
\end{figure}

\begin{table}
    \begin{subtable}{0.65\textwidth}
        \centering
    \begin{tabular}{|c|c|c|c|}
    \hline
        Data&$H$&-ELBO&NLL\\
        \hline
         8 Gaussians&  -1.92&-1.81&-1.86\\
         Checkerboard& -1&-0.73&-0.81\\
         2 Spirals& -2.34&-1.98&-2.15\\
         \hline
    \end{tabular}
    \caption{Batch Size 200.}
    \label{tab:nll}
    \end{subtable}
    \begin{subtable}{0.32\textwidth}
    \begin{tabular}{|c|c|c|}
    \hline
        Data&-ELBO&NLL\\
        \hline
        8G& -1.90&-1.91\\
        C&-0.87&-0.92\\
        2S&-2.09&-2.21\\
         \hline
    \end{tabular}
    \caption{Batch Size 40000.}
        \label{tab:nll_big}
    \end{subtable}
    \caption{Negative Log Likelihood Results on each of the 2D datasets for a Typical VAE. $H$ is the entropy of the ground truth distribution. -ELBO is the negative ELBO calculated using one sample from $\mathbf{z}\sim q_\phi(\mathbf{z}|\mathbf{x})$. NLL is the approximate negative log-likelihood. We show results for a VAE trained on a standard batch size of 200 (Table~\ref{tab:nll}) and a VAE trained on a prohibitively large batch size of 40000 (Table~\ref{tab:nll_big}).}
\end{table}

We train VAEs on several toy 2D distributions. Details of results and implementation can be found in the Jupyter notebook\footnote{Notebook can be found at https://github.com/ronaldiscool/VAETutorial}. We summarize key results below.

\subsection{Typical VAEs}

Each Typical VAE has two latent dimensions. The architecture takes roughly 1 to 2 GB of GPU and takes 20 minutes to train on a K80 on Google Colab. We trained for 60000 iterations with a batch size of 200 input samples at each iteration.

In Figure~\ref{fig:density}, we see density estimation results on several 2D datasets: a multi-modal Gaussian distribution (which we will call ``8 Gaussians``) , a uniform distribution over a checkerboard (``Checkerboard"), and a uniform distribution over 2 spirals (``2 Spirals"). A Typical VAE with 2 latent dimensions can capture the general shape of each distribution. However, the VAE also assigns non-trivial amounts of probability density to what should be low-density areas such as the space between two Gaussian clusters in ``8 Gaussians", the connection between two squares in ``Checkerboard", and the center of ``2 Spirals". As a result, less density is estimated on the ground-truth high-density areas, so in Figure~\ref{fig:density} the ground truth distributions are more yellow than the predicted distributions.

We display the negative log-likelihood of a Typical VAE trained on each dataset in Table~\ref{tab:nll}. The second column indicates the entropy $H$ of the ground truth distribution, which is also a lower bound for the negative log likelihood of a maximum likelihood model since $\mathbb{E}_{\mathbf{x} \sim p_{gt}(\mathbf{x})} -\log p_\theta(\mathbf{x}) = H + D_{KL}(p_{gt}(\mathbf{x}) || p_\theta(\mathbf{x})) \geq H$. 
The third column is the negative ELBO. The fourth column is the negative log likelihood approximated by taking 250 importance samples using Equation~\ref{eq:expectationq}. 
We can see how tight of a bound the negative ELBO is by comparing its value with the true negative log likelihood. We see that the VAE is nearly optimal on the ``8 Gaussians" dataset and the negative ELBO is almost exact. However, there is room for improvement on the other two distributions.

We visualize the correspondence between the input space and the latent space in Figure~\ref{fig:latent}. We see that the VAE copies information from the input space into the latent space and then expands the high density (colored) regions, which by extension contracts the low-density (dark) regions.
However, a significant portion of the latent space still maps to dark areas of low density, even for the ``8 Gaussians" dataset on which the VAE was nearly optimal.

Why did the VAE not achieve optimal negative log-likelihood?
Another way to look at this question is, over the course of 60000 iterations and 12 million input samples, why did the VAE fail to learn to expand the colored regions in Figure~\ref{fig:latent} until the dark regions were arbitrarily small?
Based both on our discussion in Sections~\ref{sec:prob} and~\ref{sec:information} and visual inspection of Figure~\ref{fig:latent}, it does not necessarily appear that the model is incapable of expressing a model for which the colored regions are fully expanded. 
We can partially probe the limits of our model's expressivity by training a model with a batch size of 40000. Such a model consumes 2.4 billion training samples, takes 18 hours to train, and would be prohibitively expensive in virtually any other setting besides from a synthetic 2D example. We see that there indeed is a sizable gap between what the model is capable of expressing---which is at least as powerful as the results in Figure~\ref{fig:big} and Table~\ref{tab:nll_big}---and what it actually learns given a reasonable amount of data and resources---which is reflected in the results in Figure~\ref{fig:typical} and Table~\ref{tab:nll}.
This may suggest that a large part of the problem with VAEs in this toy setting and perhaps even in practical applications may lie in optimization difficulties.

\begin{table}
    \centering
    \begin{tabular}{|c|c|c|c|}
    \hline
        Dataset&$H$&Loss&NLL\\
        \hline
         8 Gaussians&  -1.92&-1.85&-1.91\\
         Checkerboard& -1&-0.84&-0.88\\
         2 Spirals& -2.34&-2.25&-2.30\\
         \hline
    \end{tabular}
    \caption{Negative Log Likelihood Results on each of the 2D datasets for IWAEs. $H$ is the entropy of the ground truth distribution. Loss is the objective function of the IWAE, which is the approximate NLL using 10 samples. NLL is the approximate negative log-likelihood using 250 samples.}
    \label{tab:nll_iwae}
\end{table}

\begin{figure}[t]
    \centering
    \begin{subfigure}[b]{0.48\textwidth}
        \centering
                \includegraphics[width=\linewidth]{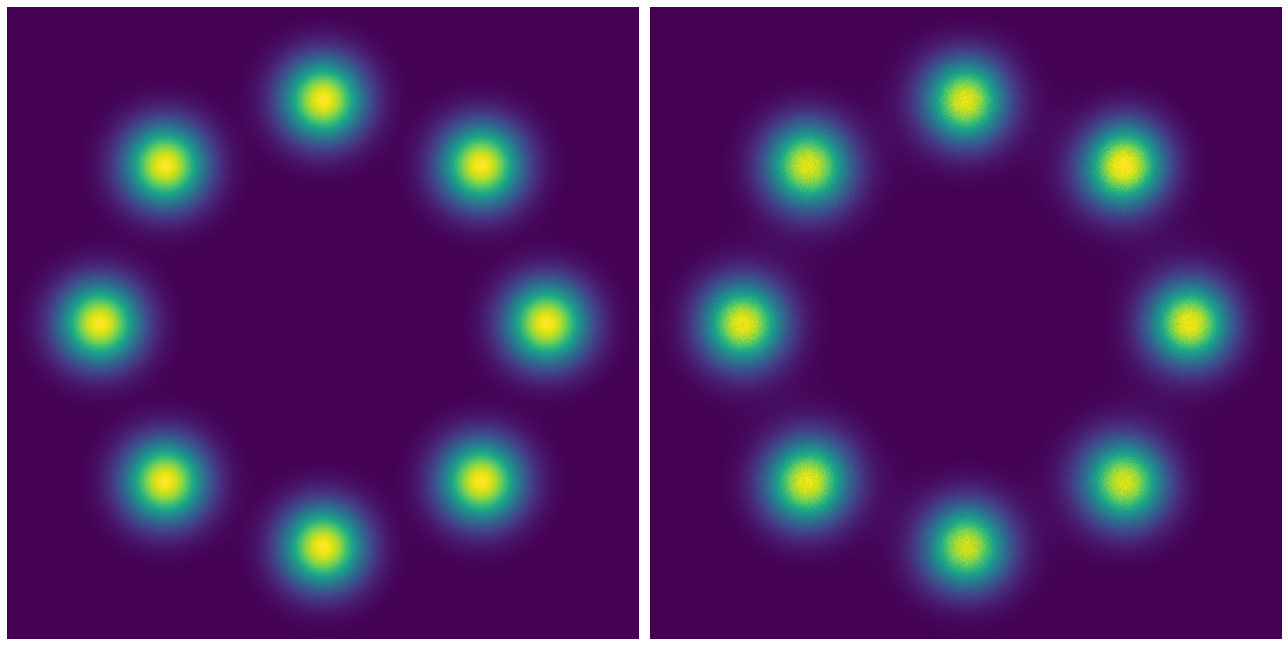}
                \includegraphics[width=\linewidth]{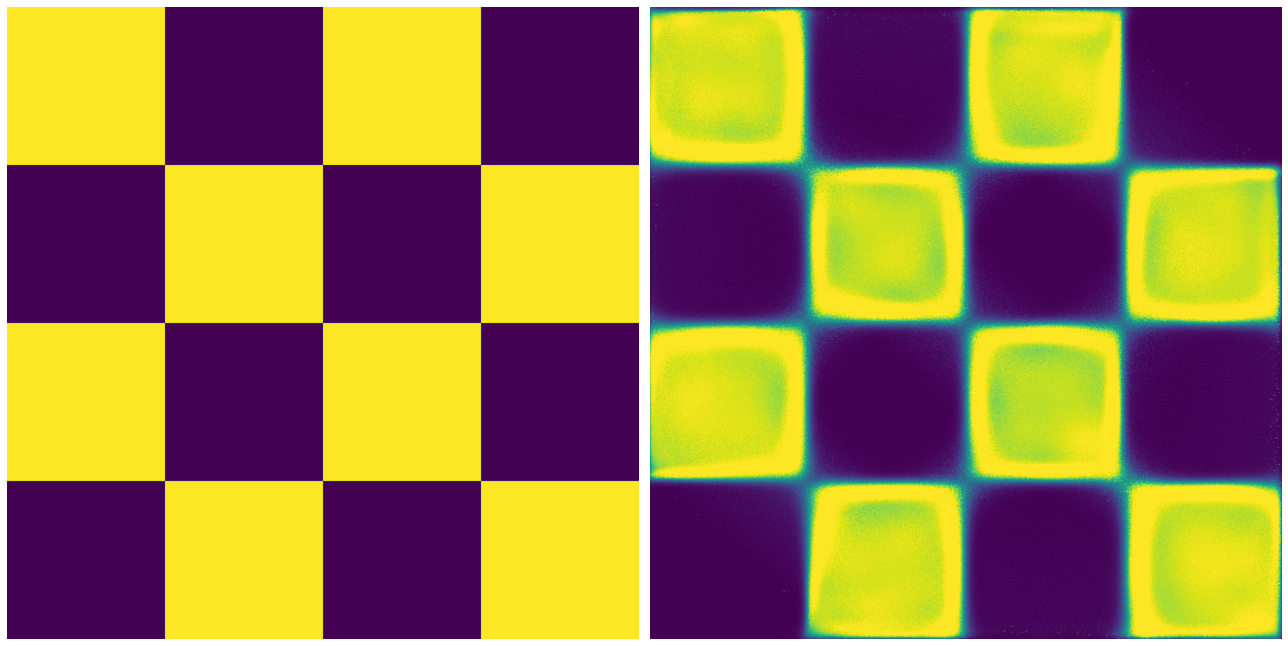}
                \includegraphics[width=\linewidth]{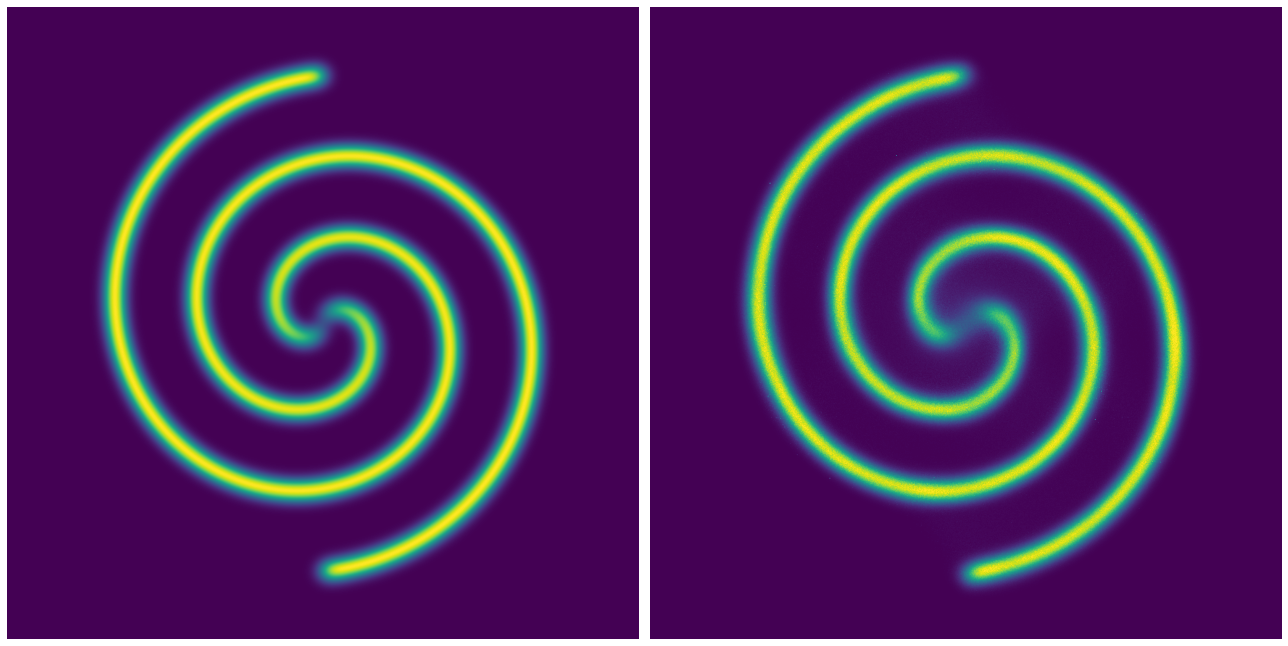}
        \caption{The ground truth probability of each pixel (left) and estimated probability from the IWAE (right).} \label{fig:density_iwae}
    \end{subfigure}
        \begin{subfigure}[b]{0.48\textwidth}
        \centering
                \includegraphics[width=\linewidth]{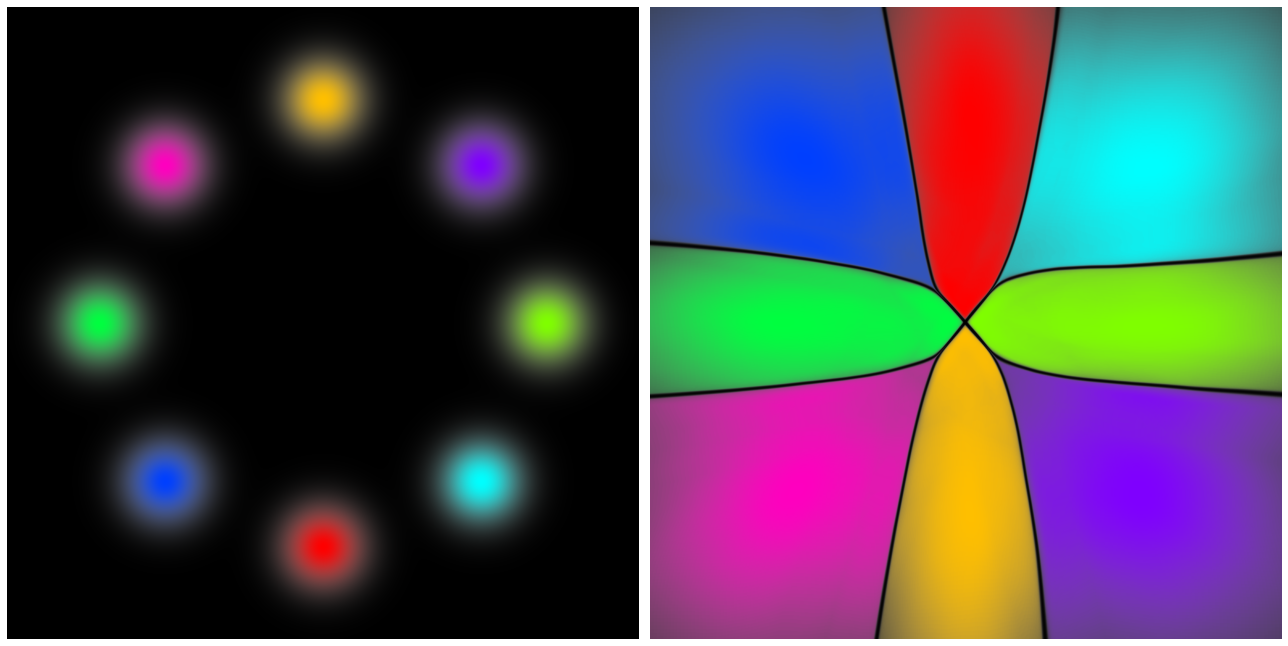}
                \includegraphics[width=\linewidth]{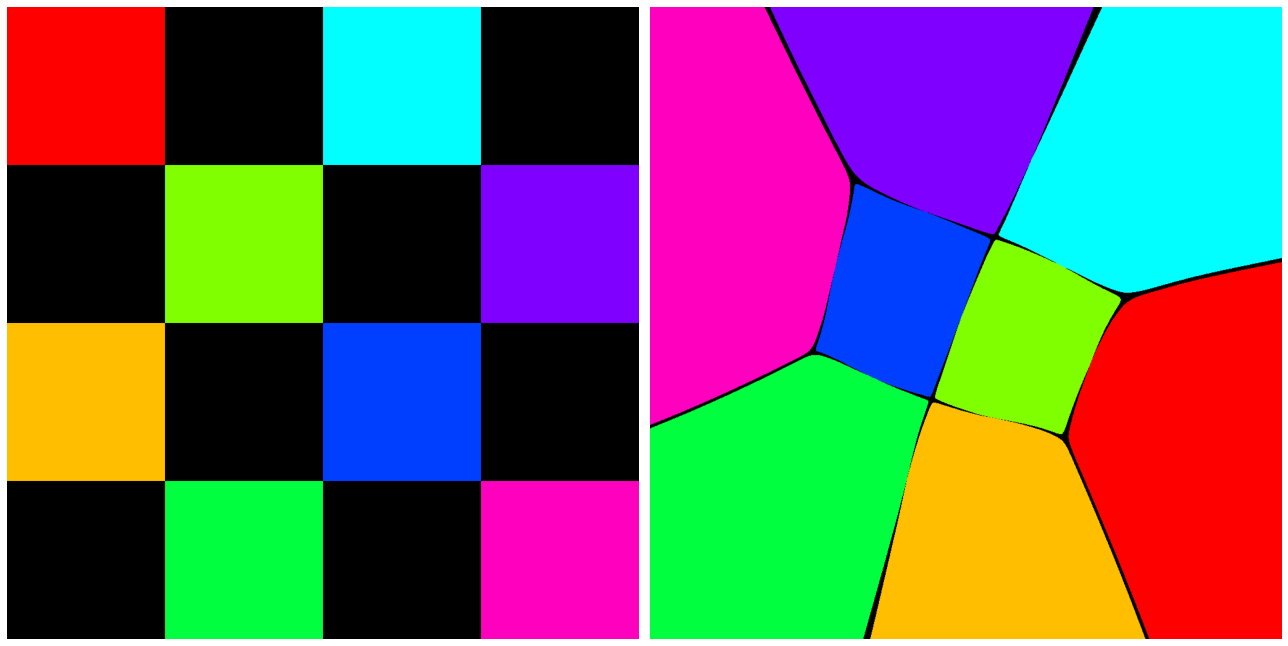}
                \includegraphics[width=\linewidth]{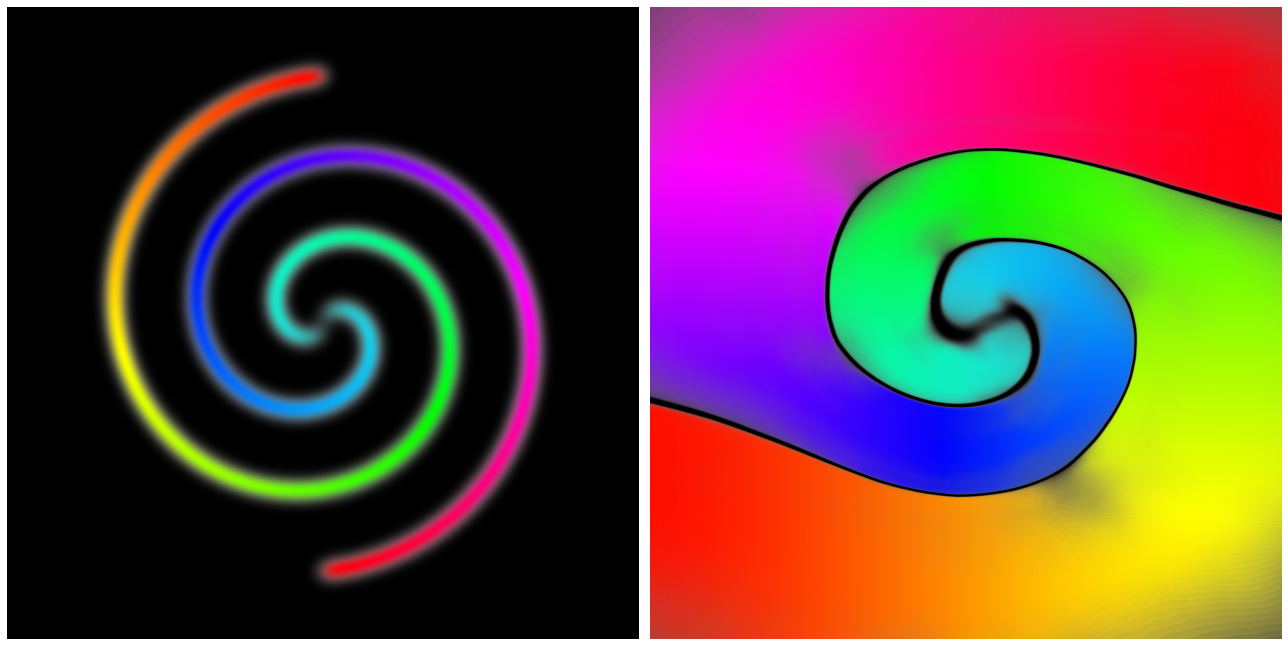}
        \caption{Color correspondence between the ground truth distribution (left) and the latent space (right).}
        \label{fig:latent_iwae}
    \end{subfigure}
    \caption{Density Estimation and Latent Space Visualization for IWAEs.}
    \label{fig:iwae}
\end{figure}

\subsection{IWAEs and Beyond}
One possible reason for the VAE's sub-optimal results may be that the decoder does not receive samples from $p_\theta(\mathbf{z})$ as input, but rather samples from $q_\phi(\mathbf{z}|\mathbf{x})$, which would presumably be centered around colored regions of the latent space.
During early phases of training, $q_\phi(\mathbf{z}|\mathbf{x})$ has high variance, allowing the decoder to encounter inputs sampled from dark regions and ``explore" the latent space, which in turn allows it to learn to expand the colored regions.
However, as the VAE grows more powerful, the regularization loss quickly increases and the variance of $q_\phi(\mathbf{z}|\mathbf{x})$ decreases. The decoder then explores increasingly fewer inputs from the dark regions, so the decoder will expand the colored region at an increasingly slower rate.
If the variance of $q_\phi(\mathbf{z}|\mathbf{x})$ decreases too quickly, then the VAE may not be able to converge to the global optimum in a reasonable amount of time.

One remedy to this limitation is the Importance-Weighted Auto-Encoder (IWAE)~\cite{burda2015importance} mentioned in Section~\ref{sec:importance}.  For its loss function, the IWAE samples from $q_\phi(\mathbf{z}|\mathbf{x})$ multiple times during each iteration of training to approximate Equation~\ref{eq:expectationq}.
One key advantage of minimizing an approximation of the negative log-likelihood instead of an upper bound is that $p_\theta(\mathbf{z}|\mathbf{x})$ no longer needs to closely match $q_\phi(\mathbf{z}|\mathbf{x})$, allowing for more flexibility.
For a visual intuition of  the utility of importance sampling, considering the following scenario: 
Suppose we have a multi-modal distribution where the chief job of the encoder is to select which cluster $\mathbf{x}$ belongs to. The VAE infers an accurate but not perfect posterior distribution $q_\phi(\mathbf{z}|\mathbf{x})$ such that $p_\theta(\mathbf{x}|\mathbf{z})=1$ with probability $0.9$ and $p_\theta(\mathbf{x}|\mathbf{z})=2^{-10}$ with probability $0.1$ (e.g. the tail end of a Gaussian extends to a region of latent space belonging to another cluster). This small chance of error is amplified in the negative ELBO, as  $-\mathbb{E}[ \log p_\theta(\mathbf{x}|\mathbf{z})] = -0.9 \times \log 1 - 0.1 \times \log 2^{-10} = -1$. On the other hand, if we swap the order of the expectation and logarithm as done in importance sampling, the accurate predictions will drown out the inaccurate ones as $- \log \mathbb{E}[ p_\theta(\mathbf{x}|\mathbf{z}) ] = -\log [0.9+0.1\times2^{-10}] = -0.15$.
Hence, compared to an IWAE, training a  VAE using the negative ELBO will encourage lower variance estimates of $q_\phi(\mathbf{z}|\mathbf{x})$, which highly discourages the VAE from exploring the dark regions in Figure~\ref{fig:latent} for which $p_\theta(\mathbf{x}|\mathbf{z})$ are low.

We train an IWAE using ten importance samples during each training iteration for 30000 iterations with a batch size of 200, which takes roughly the same amount of time as our Typical VAE but receives half the amount of input data.
We see that, compared to a VAE, the IWAE achieves better negative log-likelihood numbers in Table~\ref{tab:nll_iwae}---rivaling or even surpassing the negative log-likelihood of the VAE trained with extremely large batch size---and thus achieves a more faithful probability density map in Figure~\ref{fig:density_iwae}. 
Table~\ref{tab:nll_iwae} also shows that the objective of the IWAE is a tighter bound on $-\log p_\theta(\mathbf{x})$ compared to the negative ELBO.
However, the most striking difference can be seen in Figure~\ref{fig:latent_iwae}, in which the colored regions have been significantly expanded.
 
 From this toy example, we have seen the capabilities and limitations of a VAE in an empirical setting. While they are capable of capturing the general shape of even highly non-Gaussian distributions, the negative ELBO may prevent the VAE from efficiently filling the latent space with high-probability data-points. Even though we have shown in Sections~\ref{sec:flow} and~\ref{sec:transmission} that there are certain situations (i.e. when the latent space is fully disentangled or when the latent space is high-dimensional and the input data is discrete) where the VAE could theoretically perfectly model the ground truth input distribution, in practice these solutions could be virtually impossible to learn via gradient descent.
 
 As a result of these limitations, in addition to IWAEs, further research on VAEs include allowing for a more flexible posterior distribution using normalizing flows in which $q_\phi(\mathbf{z}|\mathbf{x})$ can accurately infer $p_\theta(\mathbf{z}|\mathbf{x})$ even when its variance is large~\cite{rezende2015variational,kingma2016improved}. 
 Other improvements to the VAE including allowing for a more flexible latent prior distribution~\cite{tomczak2018vae,bauer2019resampled} or output distribution~\cite{vlae}.
 With these improvements, VAEs are becoming an increasingly powerful probabilistic model that can be used for a variety of applications like lossless compression, generative image modeling, and representation learning.

\bibliographystyle{ieeetr}
\bibliography{bibfile}

\end{document}